\documentclass{article}

\usepackage[final]{neurips_2024}

\usepackage[utf8]{inputenc} % allow utf-8 input
\usepackage[T1]{fontenc}    % use 8-bit T1 fonts
\usepackage{hyperref}       % hyperlinks
\usepackage{url}            % simple URL typesetting
\usepackage{booktabs}       % professional-quality tables
\usepackage{amsfonts}       % blackboard math symbols
\usepackage{nicefrac}       % compact symbols for 1/2, etc.
\usepackage{microtype}      % microtypography
\usepackage{xcolor}         % colors

\usepackage{graphicx}
\usepackage{cleveref}
\usepackage{wrapfig}
\usepackage{float}
\usepackage{multirow}
\usepackage{makecell}
\usepackage{booktabs}
\usepackage{longtable}

\usepackage{tcolorbox}
\usepackage{fancyvrb}
\usepackage{geometry}
\usepackage{multicol}
\usepackage{adjustbox}
\usepackage{enumitem}
\usepackage{placeins}
\usepackage[title]{appendix}

\definecolor{customblue}{RGB}{70,130,180}    % Steel blue - nice light but visible blue
\definecolor{customred}{RGB}{178,34,34}

\title{Using LLMs for Late Multimodal Sensor Fusion for Activity Recognition}

\author{
Ilker Demirel\textsuperscript{1, 2\thanks{Work done while at Apple}},
    Karan Thakkar\textsuperscript{1,3\footnotemark[1]},
  Benjamin Elizalde\textsuperscript{1},
 Miquel Espi Marques\textsuperscript{1},\\
  \textbf{Aditya Sarathy}\textsuperscript{1},
   \textbf{Yang Bai}\textsuperscript{1},
    \textbf{Umamahesh Srinivas}\textsuperscript{1},
    \textbf{Jiajie Xu}\textsuperscript{1},
  \textbf{Shirley Ren}\textsuperscript{1}, \textbf{Jaya Narain\textsuperscript{1}}
\\
  \textsuperscript{1}Apple  \textsuperscript{2}MIT, 
 \textsuperscript{3}Johns Hopkins \\
  \ \texttt{demirel@mit.edu, jnarain@apple.com}
}

\definecolor{promptcolor}{RGB}{240, 248, 255}
\definecolor{responsecolor}{RGB}{248, 255, 240}
\definecolor{contextcolor}{RGB}{255, 250, 240}
\definecolor{bordercolor}{RGB}{100, 149, 237}

\definecolor{deepblue}{RGB}{25, 25, 112}        % Deep blue for main headers
\definecolor{crimson}{RGB}{178, 34, 34}         % Crimson for time steps
\definecolor{forestgreen}{RGB}{34, 139, 34}     % Forest green for section labels
\definecolor{darkslategray}{RGB}{47, 79, 79}    % Dark slate gray for IMU predictions
\definecolor{indigo}{RGB}{75, 0, 130}           % Indigo for additional context

\newtcolorbox{llmexample}[1]{
    colback=white,
    colframe=bordercolor,
    arc=3mm,
    boxrule=1pt,
    title={Example #1},
    fonttitle=\bfseries,
    coltitle=white,
    colbacktitle=bordercolor,
    toptitle=2mm,
    bottomtitle=2mm,
    lefttitle=5mm,
    righttitle=5mm
}

\newtcolorbox{promptbox}{
    colback=promptcolor,
    colframe=bordercolor,
    arc=2mm,
    boxrule=0.5pt,
    left=5pt,
    right=5pt,
    top=5pt,
    bottom=5pt,
    before upper={\textbf{Context in the Prompt:}\\[2pt]}
}

\newtcolorbox{responsebox}{
    colback=responsecolor,
    colframe=bordercolor,
    arc=2mm,
    boxrule=0.5pt,
    left=5pt,
    right=5pt,
    top=5pt,
    bottom=5pt,
    before upper={\textbf{Gemini-2.5-pro Response:}\\[2pt]}
}

\begin{document}

\maketitle
\begin{abstract}

Sensor data streams provide valuable information around activities and context for downstream applications, though integrating complementary information can be challenging. We show that large language models (LLMs) can be used for late fusion for activity classification from audio and motion time series data.  We curated a subset of data for diverse activity recognition across contexts (e.g., household activities, sports) from the Ego4D dataset.  Evaluated LLMs achieved 12-class zero- and one-shot classification F1-scores significantly above chance, with no task-specific training. Zero-shot classification via LLM-based fusion from modality-specific models can enable multimodal temporal applications where there is limited aligned training data for learning a shared embedding space.  Additionally, LLM-based fusion can enable model deploying without requiring additional memory and computation for targeted application-specific multimodal models.

\end{abstract}

\section{Introduction and Related Work}

%%TO-Do: 
%Emphasize flexibility of activity recognition without video data from a privacy standpoint
%% Connect to health for health workshop.  E.g., mention that multimodality and data scarcity are both common in health domains
%% Expand literature from multimodal fusion, activity classification, and LLM-time series integration literature

%% Emphasize the flexibility of not relying on video data
%% https://drive.google.com/file/d/1KrCkEWpLn0zsupdiwXFYlsZ3N9tEsKSi/view
%%
%% Comparison to video performance would be interesting
%% More data, like ego-exo 4d?

%https://ieeexplore.ieee.org/abstract/document/11123829
% https://dl.acm.org/doi/abs/10.1145/3711896.3737226
%https://arxiv.org/abs/2406.14498
%https://arxiv.org/pdf/2506.19769
%https://arxiv.org/pdf/2505.02123

%ements than modality-specific models.  

Large language models (LLM) have world knowledge and reasoning capabilities that could enable late fusion of time series modalities, allowing integration of lightweight modality-specific models without requiring additional training to learn a joint embedding space.   Here we explore reasoning capabilities of LLMs for late fusion of distinct multi-modal time series data streams for activity recognition.   We focused on time series data because its ease of recording, storage, and privacy-sensitive characteristics make time series data streams a strong practical candidate for activity recognition -- key context for many health applications.  Late fusion via LLMs enables zero-shot learning with time series data, which is particularly impactful in the health domain where there is often limited training data and privacy is critical.  Recognizing activity context can contribute to improved machine learning algorithms for health, where models may be impacted by unobserved contextual factors (e.g., interpreting heart rate data may be impacted by if a person is doing chores, socializing, or relaxing).

Extensive prior work has explored learning joint embedding spaces across modalities via a constrastive loss, and by combining constrastive and generative losses -- including CLIP (\cite{radford2021learning}), CLAP (\cite{elizalde2023clap}), and recent works in the sensor and activity recognition domains including LLaSA, Sensor2Text, and SensorLM (\cite{zheng2025review,  jiang2024neurolm, imran2024llasa, chen2024sensor2text, hong2025llm4har}).  For instance, Sensor2Text trains a Q-former fusion module while fine-tuning individual encoders to learn text representations of motion data and SensorLM employs a combination of contrastive and captioning losses to learn to describe trends in time series data.  Capturing rich temporal and complementary semantics across modalities can be challenging.  Learning a descriptive joint embedding space of time series signals requires large sample sizes and scaling to multiple modalities can be difficult because of the scarcity of aligned, paired data.  For instance, models to fuse audio, motion, and/or text data often use trained modality-specific encoders to align audio-text pairs and sensor-text pairs (\cite{zhang2025sensorlm, jiang2024neurolm, imran2024llasa}) and are limited by the availability of rich, descriptive captions for sensor data (\cite{haresamudram2025limitations}). Furthermore, such approaches do not readily poise a model to {\em integrate} information from multiple modalities when they are {\em complementary} --- in fact, contrastive approaches may actively discourage learning relevant but orthogonal information. 

% For instance, motion data indicating repetitive bending over and standing along with beeping sounds might suggest loading a laundry machine.  Common time series fusion approaches are not targeted to leverage modality-specific information in this way.

Natively multimodal models like Gemini and GPT4o can process and respond to inputs from images, video, text, and audio but typically do not directly work with sensor data (e.g., motion data).  Fine-tuning models to work with sensor data directly can enable their use for specific downstream tasks \cite{kim2024health, chow2024towards}.  For instance, HealthLLM uses prompting and fine-tuning to adapt LLMs for health prediction tasks like sleep and stress monitoring \cite{kim2024health}.  Such approaches require tailored datasets and resulting models cannot generalize to new sensor tasks.

 % ContextLLM explored sensor-based fusion for activity recognition, but included only motion-related datastreams \cite{post2025contextllm}. 

Prior work has shown that human-in-the-loop reasoning around model predictions can improve performance \cite{huang2024human}.  By training on large, diverse text corpi, large language models (LLMs) learn general world knowledge that captures how modalities relate to each other and other contexts.  Recent advances in LLMs could similarly enable agent-in-the-loop systems for diverse modalities from multiple sensors for activity classification.  Prior work has also explored late fusion via LLMs for autonomous driving and other perception tasks (\cite{ruan2025survey, hou2025driveagent}) and for motion-related data streams in activity recognition (\cite{post2025contextllm}).  
% where a model with world knowledge can combine information from multiple models and diverse modalities as well as other background information to provide high-quality contextual labels.  
To our knowledge, late fusion using LLMs for diverse multi-modal data streams for contextual activity recognition has not been previously explored in the literature.
Our contributions are: (1) the curation of a high-quality dataset for activity recognition from Ego4D, (2) benchmarking zero- and one-shot closed-set activity classification with LLMs and an initial investigation of open-ended activity classification, and (3) insights around current capabilities and future directions on strengths and weaknesses of modalities and LLM-based fusion via provided examples and ablation studies.

\section{Methods}

\begin{figure}
    \centering
    \includegraphics[width=\linewidth]{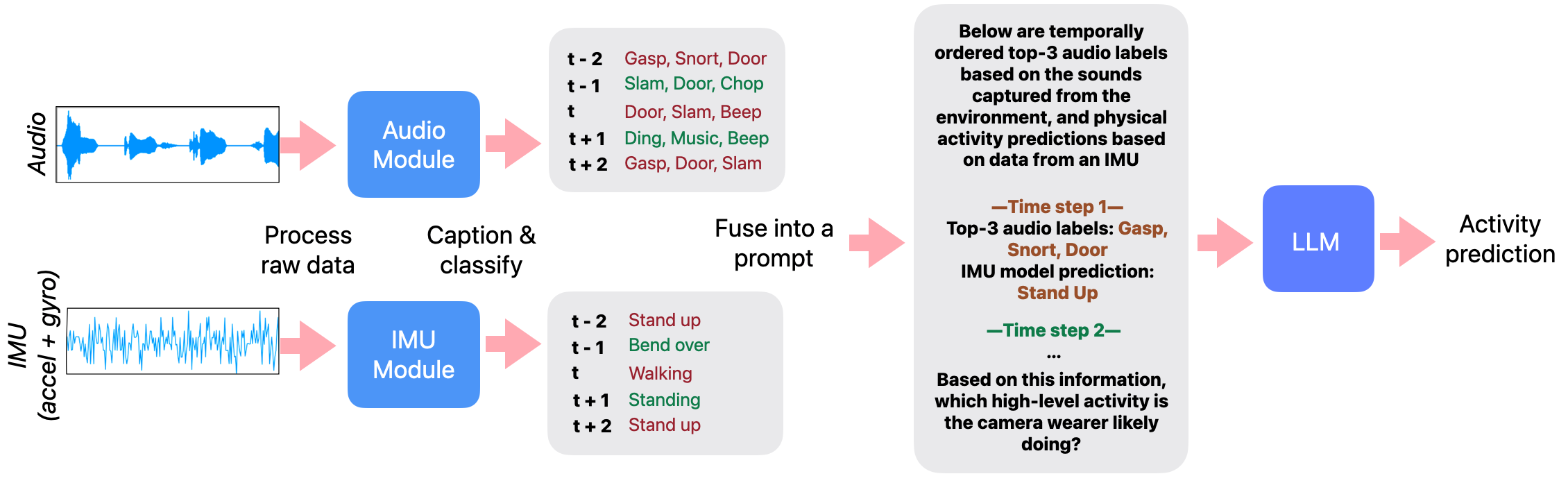}
    \caption{Model Architecture for Prompt Creation.}
    \label{fig:mod_arc}
\end{figure}

% We investigate activity classification using modality-specific models to process multi-modal information, and then use a text-only large language model to fuse information across modalities 

\subsection{Dataset}

% \begin{wraptable}{r}{0.35\textwidth}
% \centering
% \vspace{0pt}
% \caption{High-level activities}
% \vspace{5pt}
% \begin{tabular}{l}
% \toprule
% \textbf{Activity} \\
% \midrule
% Eating  \\
% Doing laundry  \\
% Cooking  \\
% Washing Dishes \\
% Vacuum Cleaning \\
% Watching TV \\
% Reading a book  \\
% Using computer \\
% Play with pets\\
% Exercise \& Workout \\
% Play basketball  \\
% Play soccer \\
% \bottomrule
% \end{tabular}
% \label{tab:activities}
% \end{wraptable}

We curated a dataset of day-to-day activities from the Ego4D dataset \cite{grauman2022ego4d} by searching for activities of daily living within the provided narrative descriptions.  The curated dataset includes 20 second samples from twelve high-level activities: vacuum cleaning, cooking, doing laundry, eating, playing basketball, playing soccer, playing with pets, reading a book, using a computer, washing dishes, watching TV, workout/weightlifting.  These activities were selected to span a range of household and fitness tasks, and based on their prevalence in the larger dataset.  The timestamp and video IDs of the segments used in our analysis are provided in the supplemental material, to enable replication of this task by others (see \Cref{tab:ego4d_metadata}).  While the evaluated dataset is small, it is high-quality -- a high proportion of the larger dataset did not have IMU or audio data, and information-rich segments for contextual activities are sparsely distributed throughout the dataset.  In the analyzed subset, each segment was manually validated to ensure it contained a representative sample of the labeled activity.  To our knowledge, it is one of the only readily curated sets of multi-modal audio and motion data for assessing diverse contextual activity recognition.

\subsection{Modeling}

The exploration includes the following modalities:
\begin{itemize}[nosep]

    \item \textbf{Audio data}, processed using MS CLAP (\cite{elizalde2023clap}) to generate audio captions and VGGish to generate audio labels \cite{hershey2017cnn}
    \item \textbf{Motion data} 3-axis accelerometer and 3-axis gyroscope from an inertial measurement unit (IMU), processed using an activity classification model trained on the target dataset
    \item \textbf{Extra Context (Synthetic)} Additional descriptive information like setting (e.g., indoor/outdoor) and heart rate zone, that could be captured from sensors.  For our experiments, we synthetically generated this information in order to evaluate its utility in modality integration
\end{itemize}

Audio data was processed in 2 second windows and IMU data in 4 second windows. IMU windows had 2 seconds overlaps and the centers of IMU and audio windows were aligned. Experiments utilized two Large Language Models (LLMs): a large, state-of-the-art model, Gemini-2.5-pro, and a smaller, high-performance model, DeepSeek-R1-Distill-Qwen-32B.  Language models were prompted to classify the contextual activity given per-modality predictions for each time step (Figure \ref{fig:mod_arc}).  The full prompt and one-shot example are in Appendix \ref{sec:prompts}.  Ablation studies were conducted to investigate the contribution of each modality, different user contexts, and the integration capabilities of the LLMs. 

\textbf{Closed-set evaluation} We evaluated performance of zero- and one-shot classification in a closed-set task.  The list of high-level activities was appended at the end of the prompt (Appendix \ref{sec:prompts}) and the model was instructed to select the most likely activity given the information in the prompt. We reported accuracy and macro-F1 score. The zero-shot setting did not include any examples in the prompt and the one-shot setting included a single illustrative example with included reasoning.

\textbf{Open-ended evaluation} A pre-defined list of possible activities may not always be available.  To assess model performance in a more challenging open-ended scenario, we evaluated performance without including predefined list of candidate activities. Results included sample outputs for qualitative discussion and a quantitative metric where \texttt{Qwen-32B} mapped the initial, unconstrained activity prediction to the best-fitting label from our original closed set.  This evaluation has limitations since it introduces a second inference step and does not directly evaluate the quality of the open-ended output, but it provided an easily measurable metric around how close the open-ended response was to the actual class and enabled understanding comparative performances.

\section{Results and Discussion}

% Here, we present results for the performance of our approach in determining the ``high-level activity'' from audio and IMU signals, and possibly some additional user context (such as ``user is indoors''). We perform ablation studies to investigate the contribution of each modality, different user contexts, and the capability of the LLM in the final step to the overall performance. 

The results for the one-shot closed-set evaluation are in \Cref{tab:closed_set_results_oneshot}.  Zero-shot results (which had similar trends) are provided in Appendix Tables 3 and 4.  The results of the open-ended inference task are summarized in \Cref{tab:open_ended_results_oneshot}.  95\% confidence intervals were calculated via bootstrapping.  Appendix \ref{sec: per-activity} shows per-activity results for fusion with each LLM for each ablation and Appendix \ref{sec:examples} shows a selection of example temporal inputs and reasoning and reasoning responses.

Both models achieved classification results well above chance (or 8.3\% accuracy) given audio captions, labels, and activity predictions.   Among the modality-specific models included, the audio caption predictions from CLAP were most informative towards overall activity predictions.  Prediction accuracy using audio labels was impacted by inaccurate class predictions that heavily influenced reasoning (as in the second example in Appendix \ref{sec:examples}, where the prediction of animal-related sounds incorrectly drove the model prediction).  Prediction accuracy using activity labels was similarly influenced by mispredictions by the modality specific-model -- for instance, when minimal motion was recognized from recorded data, the model tended to predict stationary activities like 'eating' or 'reading a book'.  This highlights the importance of both accurate modality-specific models and flexible LLM reasoning capabilities that don't overfit to anomalous predictions. 
The inclusion of synthetic context, which was designed to mimic information that could be interpreted or collected from sensors, improved performance compared to the predictions that used audio and IMU data only.  Including additional sensor-based information in activity recognition can improve performance, even in simple aggregate forms like heart rate ranges.

\renewcommand{\arraystretch}{0.8}
\begin{table}[h]
\centering
\caption{Closed-set high-level activity classification performance in the one-shot setting ($95\%$ CIs).}
\label{tab:closed_set_results_oneshot}
\setlength{\tabcolsep}{4pt} % Adjust this value to control column spacing (default is 6pt)
\begin{tabular}{lcccccc}
\toprule
\multirow{2}{*}{\makecell{\textbf{Model}\\\textbf{(Setting)}}} & \multicolumn{4}{c}{\textbf{Modalities}} & \multirow{2}{*}{\textbf{Accuracy (\%)}} & \multirow{2}{*}{\textbf{Macro-F1 (\%)}} \\
\cmidrule(lr){2-5}
& \makecell{\textbf{Audio}\\\textbf{Caption}} & \makecell{\textbf{Audio}\\\textbf{Labels}} & \makecell{\textbf{IMU Activity}\\\textbf{Pred.}} & \makecell{\textbf{Extra}\\\textbf{Context}} & & \\
\midrule
\multirow{5}{*}{\makecell{\texttt{ Gemini-}\\\texttt{2.5-pro}\\(closed-set)}} 
& \textcolor{customblue}{\checkmark} & \textcolor{customblue}{\checkmark} & \textcolor{customblue}{\checkmark} & \textcolor{customblue}{\checkmark} & 68 (59, 76) & 66 (56, 74) \\
& \textcolor{customblue}{\checkmark} & \textcolor{customblue}{\checkmark} & \textcolor{customblue}{\checkmark} & \textcolor{customred}{\texttimes} & 63 (55, 72) & 62 (52, 70) \\

& \textcolor{customblue}{\checkmark} &  \textcolor{customred}{\texttimes} & \textcolor{customred}{\texttimes} & \textcolor{customred}{\texttimes} & 60 (50, 69) & 60 (50, 67) \\
& \textcolor{customred}{\texttimes} & \textcolor{customblue}{\checkmark} &  \textcolor{customred}{\texttimes} & \textcolor{customred}{\texttimes} & 41 (32, 50) & 35 (26, 43) \\
& \textcolor{customred}{\texttimes} & \textcolor{customred}{\texttimes}  & \textcolor{customblue}{\checkmark} & \textcolor{customred}{\texttimes} & 13 (8, 20) & 10 (5, 14) \\
\midrule
\multirow{5}{*}{\makecell{\texttt{Qwen-32B}\\(closed-set)}} 
& \textcolor{customblue}{\checkmark} & \textcolor{customblue}{\checkmark} & \textcolor{customblue}{\checkmark} & \textcolor{customblue}{\checkmark} & 56 (47, 65) & 56 (45, 64) \\
& \textcolor{customblue}{\checkmark} & \textcolor{customblue}{\checkmark} & \textcolor{customblue}{\checkmark} & \textcolor{customred}{\texttimes} & 51 (43, 60) & 51 (41, 57) \\
& \textcolor{customblue}{\checkmark} &  \textcolor{customred}{\texttimes} & \textcolor{customred}{\texttimes} & \textcolor{customred}{\texttimes} & 52 (43, 61) & 52 (42, 60) \\
& \textcolor{customred}{\texttimes} & \textcolor{customblue}{\checkmark} &  \textcolor{customred}{\texttimes} & \textcolor{customred}{\texttimes} & 42 (33, 51) & 36 (29, 42) \\
& \textcolor{customred}{\texttimes} & \textcolor{customred}{\texttimes}  & \textcolor{customblue}{\checkmark} & \textcolor{customred}{\texttimes} & 13 (8, 20) & 8 (3, 12) \\
\bottomrule
\end{tabular}
\end{table}

\renewcommand{\arraystretch}{0.8}
\begin{table}[h]
\centering
\caption{Open-ended high-level activity classification performance in the one-shot setting ($95\%$ CIs).}
\label{tab:open_ended_results_oneshot}
\setlength{\tabcolsep}{4pt} % Adjust this value to control column spacing (default is 6pt)
\begin{tabular}{lcccccc}
\toprule
\multirow{2}{*}{\makecell{\textbf{Model}\\\textbf{(Setting)}}} & \multicolumn{4}{c}{\textbf{Modalities}} & \multirow{2}{*}{\textbf{Accuracy (\%)}} & \multirow{2}{*}{\textbf{Macro-F1 (\%)}} \\
\cmidrule(lr){2-5}
& \makecell{\textbf{Audio}\\\textbf{Caption}} & \makecell{\textbf{Audio}\\\textbf{Labels}} & \makecell{\textbf{IMU Activity}\\\textbf{Pred.}} & \makecell{\textbf{Extra}\\\textbf{Context}} & & \\

\midrule
\multirow{5}{*}{\makecell{\texttt{Gemini-}\\\texttt{2.5-pro}\\(open-ended)}} 
& \textcolor{customblue}{\checkmark} & \textcolor{customblue}{\checkmark} & \textcolor{customblue}{\checkmark} & \textcolor{customblue}{\checkmark} & 58 (50, 68) & 58 (49, 65) \\
& \textcolor{customblue}{\checkmark} & \textcolor{customblue}{\checkmark} & \textcolor{customblue}{\checkmark} & \textcolor{customred}{\texttimes} & 54 (45, 63) & 52 (42, 60) \\

& \textcolor{customblue}{\checkmark} &  \textcolor{customred}{\texttimes} & \textcolor{customred}{\texttimes} & \textcolor{customred}{\texttimes} & 53 (44, 63) & 51 (42, 59) \\
& \textcolor{customred}{\texttimes} & \textcolor{customblue}{\checkmark} &  \textcolor{customred}{\texttimes} & \textcolor{customred}{\texttimes} & 39 (31, 48) & 35 (26, 42) \\
& \textcolor{customred}{\texttimes} & \textcolor{customred}{\texttimes}  & \textcolor{customblue}{\checkmark} & \textcolor{customred}{\texttimes} & 8 (4, 13) & 5 (2, 9) \\

\midrule
\multirow{5}{*}{\makecell{\texttt{Qwen-32B}\\(open-ended)}} 
& \textcolor{customblue}{\checkmark} & \textcolor{customblue}{\checkmark} & \textcolor{customblue}{\checkmark} & \textcolor{customblue}{\checkmark} & 51 (43, 59) & 53 (43, 60) \\
& \textcolor{customblue}{\checkmark} & \textcolor{customblue}{\checkmark} & \textcolor{customblue}{\checkmark} & \textcolor{customred}{\texttimes} & 46 (37, 55) & 46 (36, 53) \\
& \textcolor{customblue}{\checkmark} &  \textcolor{customred}{\texttimes} & \textcolor{customred}{\texttimes} & \textcolor{customred}{\texttimes} & 53 (43, 61) & 54 (45, 61) \\
& \textcolor{customred}{\texttimes} & \textcolor{customblue}{\checkmark} &  \textcolor{customred}{\texttimes} & \textcolor{customred}{\texttimes} & 34 (26, 43) & 33 (25, 40) \\
& \textcolor{customred}{\texttimes} & \textcolor{customred}{\texttimes}  & \textcolor{customblue}{\checkmark} & \textcolor{customred}{\texttimes} & 13 (7, 19) & 8 (3, 13) \\
\bottomrule
\end{tabular}
\end{table}

While the IMU activity predictions did not necessarily improve aggregate performance, they were used by the model in reasoning correct predictions in some cases (as in the first example in Appendix \ref{sec:examples}) and for some classes (e.g., high motion classes like 'playing soccer' as shown in Tables \ref{tab:per_task_ablation_qwen} and \ref{tab:per_task_ablation_gemini} in Appendix \ref{sec: per-activity}).  The IMU classification model used in these experiments had a limited output space compared to the audio models.  A more powerful and flexible motion-based model would likely further help disambiguate between high- and low- motion activities.  Performance was also significantly above chance in the open-ended evaluation experiments where no activity list was provided (\Cref{tab:open_ended_results_oneshot}), highlighting the promise of using LLM-based reasoning for modality fusion in difficult tasks without requiring additional training for modality alignment.   The third and fourth examples in Appendix \ref{sec:examples} show the model's reasoning for inferring the context of cooking in open-ended cases, and the fifth example shows an example where the model incorrectly predicts playing tennis instead of basketball, based on the motion predictions and recognized ball sounds.

\section{Conclusions and Future Work}
We presented a first analysis of LLM-based fusion for activity recognition from time series data, showing that LLMs be used to fuse predictions from modality-specific models for activity recognition without requiring additional training or modality alignment. This is particularly salient for data-scarce time series tasks in sensing and health.  Sensor-based activity recognition also has the potential to enhance the deployment of health models by providing valuable context for predictions, and systems can be easily extended to include additional modalities.  We include a curated list of segments from a public multimodal dataset, Ego4D, to enable replicating and building upon this work.  The presented experiments included a limited number of samples, activities, and a limited output space for the IMU modality.  Future work will expand on the evaluated data and modality-specific models, and also explore strategies for training LLMs for targeted reasoning skills (\cite{muennighoff2025s1, yue2024dots, hojer2025improving}) for modality integration.

% Add more examples following the same pattern...

% \end{document}

\bibliographystyle{unsrtnat}
\bibliography{ref}

%%%%%%%%%%%%%%%%%%%%%%%%%%%%%%%%%%%%%%%%%%%%%%%%%%%%%%%%%%%%
\clearpage
\appendix

\begin{appendices}

\section{Zero-shot results for closed-set classification}

\begin{table}[h!]
\caption{Closed-set high-level activity classification performance in the zero-shot setting ($95\%$ CIs calculated with bootstrapping).}
\label{tab:closed_set_results_zeroshot_app}
\begin{tabular}{lcccccc}
\toprule
\multirow{2}{*}{\textbf{Model}} & \multicolumn{4}{c}
{\textbf{Modalities}} & \multirow{2}{*}{\textbf{Accuracy (\%)}} & \multirow{2}{*}{\textbf{Macro-F1 (\%)}} \\
\cmidrule(lr){2-5}
& \makecell{\textbf{Audio}\\\textbf{Caption}} & \makecell{\textbf{Audio}\\\textbf{Labels}} & \makecell{\textbf{IMU Activity}\\\textbf{Pred.}} & \makecell{\textbf{Extra}\\\textbf{Context}} & & \\
\midrule
\multirow{5}{*}{\makecell{\texttt{ Gemini-}\\\texttt{2.5-pro}}} 
& \textcolor{customblue}{\checkmark} & \textcolor{customblue}{\checkmark} & \textcolor{customblue}{\checkmark} & \textcolor{customblue}{\checkmark} & 63 (55, 72) & 61 (52, 68) \\
& \textcolor{customblue}{\checkmark} & \textcolor{customblue}{\checkmark} & \textcolor{customblue}{\checkmark} & \textcolor{customred}{\texttimes} & 60 (52, 69) & 59 (49, 66) \\

& \textcolor{customblue}{\checkmark} &  \textcolor{customred}{\texttimes} & \textcolor{customred}{\texttimes} & \textcolor{customred}{\texttimes} & 60 (51, 68) & 59 (49, 66) \\
& \textcolor{customred}{\texttimes} & \textcolor{customblue}{\checkmark} &  \textcolor{customred}{\texttimes} & \textcolor{customred}{\texttimes} & 49 (41, 58) & 45 (36, 52) \\
& \textcolor{customred}{\texttimes} & \textcolor{customred}{\texttimes}  & \textcolor{customblue}{\checkmark} & \textcolor{customred}{\texttimes} & 11 (6, 17) & 7 (3, 10) \\

\midrule
\multirow{5}{*}{\makecell{\texttt{Qwen-32B}}} 
& \textcolor{customblue}{\checkmark} & \textcolor{customblue}{\checkmark} & \textcolor{customblue}{\checkmark} & \textcolor{customblue}{\checkmark} & 52 (43, 61) & 52 (41, 60) \\
& \textcolor{customblue}{\checkmark} & \textcolor{customblue}{\checkmark} & \textcolor{customblue}{\checkmark} & \textcolor{customred}{\texttimes} & 48 (40, 58) & 47 (37, 55) \\

& \textcolor{customblue}{\checkmark} &  \textcolor{customred}{\texttimes} & \textcolor{customred}{\texttimes} & \textcolor{customred}{\texttimes} & 48 (40, 58) & 48 (39, 56) \\
& \textcolor{customred}{\texttimes} & \textcolor{customblue}{\checkmark} &  \textcolor{customred}{\texttimes} & \textcolor{customred}{\texttimes} & 41 (33, 50) & 38 (29, 45) \\
& \textcolor{customred}{\texttimes} & \textcolor{customred}{\texttimes}  & \textcolor{customblue}{\checkmark} & \textcolor{customred}{\texttimes} & 10 (5, 15) & 4 (1, 6) \\
\bottomrule
\end{tabular}
\end{table}

\section{Per-activity performance for closed-set classification}
\label{sec: per-activity}
\FloatBarrier
\clearpage

\begin{table}[t!]
%\centering
\caption{Closed-set one-shot per-activity performance with \texttt{ Gemini-2.5-pro}.}
\label{tab:per_task_ablation_gemini}
\footnotesize
\begin{tabular}{lcccccccc}
\toprule
\multirow{3}{*}{\makecell{\textbf{High-level}\\\textbf{Activity}}} & \multicolumn{4}{c}{\textbf{Modalities}} & \multirow{3}{*}{\textbf{Precision}} & \multirow{3}{*}{\textbf{Recall}}& \multirow{3}{*}{\textbf{F1}} \\
\cmidrule(lr){2-5}
& \makecell{\textbf{Audio}\\\textbf{Caption}} & \makecell{\textbf{Audio}\\\textbf{Labels}} & \makecell{\textbf{IMU Activity}\\\textbf{Pred.}} & \makecell{\textbf{Extra}\\\textbf{Context}} &  \\
\midrule

\multirow{8}{*}{\makecell[l]{Vacuum\\Cleaning}} 
& \textcolor{customblue}{\checkmark} & \textcolor{customblue}{\checkmark} & \textcolor{customblue}{\checkmark} & \textcolor{customblue}{\checkmark} & 0.50 & 0.40 & 0.44 \\
& \textcolor{customblue}{\checkmark} & \textcolor{customblue}{\checkmark} & \textcolor{customblue}{\checkmark} & \textcolor{customred}{\texttimes} & 0.38 & 0.30 & 0.33 \\

& \textcolor{customblue}{\checkmark} & \textcolor{customred}{\texttimes} & \textcolor{customred}{\texttimes} & \textcolor{customred}{\texttimes} & 0.33 & 0.30 & 0.32 \\
& \textcolor{customred}{\texttimes} & \textcolor{customblue}{\checkmark} & \textcolor{customred}{\texttimes} & \textcolor{customred}{\texttimes} & 1.00 & 0.20 & 0.33 \\
& \textcolor{customred}{\texttimes} & \textcolor{customred}{\texttimes} & \textcolor{customblue}{\checkmark} & \textcolor{customred}{\texttimes} & 0.20 & 0.10 & 0.13 \\
\midrule

\multirow{8}{*}{Cooking} 
& \textcolor{customblue}{\checkmark} & \textcolor{customblue}{\checkmark} & \textcolor{customblue}{\checkmark} & \textcolor{customblue}{\checkmark} & 0.69 & 0.90 & 0.78 \\
& \textcolor{customblue}{\checkmark} & \textcolor{customblue}{\checkmark} & \textcolor{customblue}{\checkmark} & \textcolor{customred}{\texttimes} & 0.67 & 1.00 & 0.80 \\

& \textcolor{customblue}{\checkmark} & \textcolor{customred}{\texttimes} & \textcolor{customred}{\texttimes} & \textcolor{customred}{\texttimes} & 0.64 & 0.70 & 0.67 \\
& \textcolor{customred}{\texttimes} & \textcolor{customblue}{\checkmark} & \textcolor{customred}{\texttimes} & \textcolor{customred}{\texttimes} & 0.47 & 0.90 & 0.62 \\
& \textcolor{customred}{\texttimes} & \textcolor{customred}{\texttimes} & \textcolor{customblue}{\checkmark} & \textcolor{customred}{\texttimes} & 0.00 & 0.00 & 0.00 \\
\midrule

\multirow{8}{*}{\makecell[l]{Doing\\laundry}} 
& \textcolor{customblue}{\checkmark} & \textcolor{customblue}{\checkmark} & \textcolor{customblue}{\checkmark} & \textcolor{customblue}{\checkmark} & 0.80 & 0.40 & 0.53 \\
& \textcolor{customblue}{\checkmark} & \textcolor{customblue}{\checkmark} & \textcolor{customblue}{\checkmark} & \textcolor{customred}{\texttimes} & 0.80 & 0.40 & 0.53 \\
& \textcolor{customblue}{\checkmark} & \textcolor{customred}{\texttimes} & \textcolor{customred}{\texttimes} & \textcolor{customred}{\texttimes} & 0.62 & 0.50 & 0.56 \\
& \textcolor{customred}{\texttimes} & \textcolor{customblue}{\checkmark} & \textcolor{customred}{\texttimes} & \textcolor{customred}{\texttimes} & 0.00 & 0.00 & 0.00 \\
& \textcolor{customred}{\texttimes} & \textcolor{customred}{\texttimes} & \textcolor{customblue}{\checkmark} & \textcolor{customred}{\texttimes} & 0.20 & 0.30 & 0.24 \\
\midrule

\multirow{8}{*}{Eating} 
& \textcolor{customblue}{\checkmark} & \textcolor{customblue}{\checkmark} & \textcolor{customblue}{\checkmark} & \textcolor{customblue}{\checkmark} & 0.86 & 0.60 & 0.71 \\
& \textcolor{customblue}{\checkmark} & \textcolor{customblue}{\checkmark} & \textcolor{customblue}{\checkmark} & \textcolor{customred}{\texttimes} & 0.88 & 0.70 & 0.78 \\
& \textcolor{customblue}{\checkmark} & \textcolor{customred}{\texttimes} & \textcolor{customred}{\texttimes} & \textcolor{customred}{\texttimes} & 1.00 & 0.70 & 0.82 \\
& \textcolor{customred}{\texttimes} & \textcolor{customblue}{\checkmark} & \textcolor{customred}{\texttimes} & \textcolor{customred}{\texttimes} & 0.39 & 0.70 & 0.50 \\
& \textcolor{customred}{\texttimes} & \textcolor{customred}{\texttimes} & \textcolor{customblue}{\checkmark} & \textcolor{customred}{\texttimes} & 0.00 & 0.00 & 0.00 \\
\midrule

\multirow{8}{*}{\makecell[l]{Playing\\basketball}} 
& \textcolor{customblue}{\checkmark} & \textcolor{customblue}{\checkmark} & \textcolor{customblue}{\checkmark} & \textcolor{customblue}{\checkmark} & 0.75 & 0.90 & 0.82 \\
& \textcolor{customblue}{\checkmark} & \textcolor{customblue}{\checkmark} & \textcolor{customblue}{\checkmark} & \textcolor{customred}{\texttimes} & 0.70 & 0.70 & 0.70 \\
& \textcolor{customblue}{\checkmark} & \textcolor{customred}{\texttimes} & \textcolor{customred}{\texttimes} & \textcolor{customred}{\texttimes} & 0.82 & 0.90 & 0.86 \\
& \textcolor{customred}{\texttimes} & \textcolor{customblue}{\checkmark} & \textcolor{customred}{\texttimes} & \textcolor{customred}{\texttimes} & 1.00 & 0.40 & 0.57 \\
& \textcolor{customred}{\texttimes} & \textcolor{customred}{\texttimes} & \textcolor{customblue}{\checkmark} & \textcolor{customred}{\texttimes} & 0.00 & 0.00 & 0.00 \\
\midrule

\multirow{8}{*}{\makecell[l]{Playing\\soccer}} 
& \textcolor{customblue}{\checkmark} & \textcolor{customblue}{\checkmark} & \textcolor{customblue}{\checkmark} & \textcolor{customblue}{\checkmark} & 1.00 & 0.40 & 0.57 \\
& \textcolor{customblue}{\checkmark} & \textcolor{customblue}{\checkmark} & \textcolor{customblue}{\checkmark} & \textcolor{customred}{\texttimes} & 1.00 & 0.30 & 0.46 \\
& \textcolor{customblue}{\checkmark} & \textcolor{customred}{\texttimes} & \textcolor{customred}{\texttimes} & \textcolor{customred}{\texttimes} & 0.80 & 0.40 & 0.53 \\
& \textcolor{customred}{\texttimes} & \textcolor{customblue}{\checkmark} & \textcolor{customred}{\texttimes} & \textcolor{customred}{\texttimes} & 0.00 & 0.00 & 0.00 \\
& \textcolor{customred}{\texttimes} & \textcolor{customred}{\texttimes} & \textcolor{customblue}{\checkmark} & \textcolor{customred}{\texttimes} & 0.67 & 0.40 & 0.50 \\
\midrule

\multirow{8}{*}{\makecell[l]{Playing\\with pets}} 
& \textcolor{customblue}{\checkmark} & \textcolor{customblue}{\checkmark} & \textcolor{customblue}{\checkmark} & \textcolor{customblue}{\checkmark} & 0.53 & 1.00 & 0.69 \\
& \textcolor{customblue}{\checkmark} & \textcolor{customblue}{\checkmark} & \textcolor{customblue}{\checkmark} & \textcolor{customred}{\texttimes} & 0.56 & 1.00 & 0.71 \\

& \textcolor{customblue}{\checkmark} & \textcolor{customred}{\texttimes} & \textcolor{customred}{\texttimes} & \textcolor{customred}{\texttimes} & 0.71 & 1.00 & 0.83 \\
& \textcolor{customred}{\texttimes} & \textcolor{customblue}{\checkmark} & \textcolor{customred}{\texttimes} & \textcolor{customred}{\texttimes} & 0.38 & 0.80 & 0.52 \\
& \textcolor{customred}{\texttimes} & \textcolor{customred}{\texttimes} & \textcolor{customblue}{\checkmark} & \textcolor{customred}{\texttimes} & 0.00 & 0.00 & 0.00 \\
\midrule

\multirow{8}{*}{\makecell[l]{Reading\\a book}} 
& \textcolor{customblue}{\checkmark} & \textcolor{customblue}{\checkmark} & \textcolor{customblue}{\checkmark} & \textcolor{customblue}{\checkmark} & 1.00 & 0.30 & 0.46 \\
& \textcolor{customblue}{\checkmark} & \textcolor{customblue}{\checkmark} & \textcolor{customblue}{\checkmark} & \textcolor{customred}{\texttimes} & 1.00 & 0.30 & 0.46 \\

& \textcolor{customblue}{\checkmark} & \textcolor{customred}{\texttimes} & \textcolor{customred}{\texttimes} & \textcolor{customred}{\texttimes} & 0.75 & 0.30 & 0.43 \\
& \textcolor{customred}{\texttimes} & \textcolor{customblue}{\checkmark} & \textcolor{customred}{\texttimes} & \textcolor{customred}{\texttimes} & 0.67 & 0.20 & 0.31 \\
& \textcolor{customred}{\texttimes} & \textcolor{customred}{\texttimes} & \textcolor{customblue}{\checkmark} & \textcolor{customred}{\texttimes} & 0.50 & 0.10 & 0.17 \\
\midrule

\multirow{8}{*}{\makecell[l]{Using\\computer}} 
& \textcolor{customblue}{\checkmark} & \textcolor{customblue}{\checkmark} & \textcolor{customblue}{\checkmark} & \textcolor{customblue}{\checkmark} & 0.67 & 0.80 & 0.73 \\
& \textcolor{customblue}{\checkmark} & \textcolor{customblue}{\checkmark} & \textcolor{customblue}{\checkmark} & \textcolor{customred}{\texttimes} & 0.73 & 0.80 & 0.76 \\
& \textcolor{customblue}{\checkmark} & \textcolor{customred}{\texttimes} & \textcolor{customred}{\texttimes} & \textcolor{customred}{\texttimes} & 0.64 & 0.70 & 0.67 \\
& \textcolor{customred}{\texttimes} & \textcolor{customblue}{\checkmark} & \textcolor{customred}{\texttimes} & \textcolor{customred}{\texttimes} & 0.62 & 0.80 & 0.70 \\
& \textcolor{customred}{\texttimes} & \textcolor{customred}{\texttimes} & \textcolor{customblue}{\checkmark} & \textcolor{customred}{\texttimes} & 0.00 & 0.00 & 0.00 \\
\midrule

\multirow{8}{*}{\makecell[l]{Washing\\dishes}} 
& \textcolor{customblue}{\checkmark} & \textcolor{customblue}{\checkmark} & \textcolor{customblue}{\checkmark} & \textcolor{customblue}{\checkmark} & 0.88 & 0.70 & 0.78 \\
& \textcolor{customblue}{\checkmark} & \textcolor{customblue}{\checkmark} & \textcolor{customblue}{\checkmark} & \textcolor{customred}{\texttimes} & 0.88 & 0.70 & 0.78 \\
& \textcolor{customblue}{\checkmark} & \textcolor{customred}{\texttimes} & \textcolor{customred}{\texttimes} & \textcolor{customred}{\texttimes} & 0.86 & 0.60 & 0.71 \\
& \textcolor{customred}{\texttimes} & \textcolor{customblue}{\checkmark} & \textcolor{customred}{\texttimes} & \textcolor{customred}{\texttimes} & 1.00 & 0.10 & 0.18 \\
& \textcolor{customred}{\texttimes} & \textcolor{customred}{\texttimes} & \textcolor{customblue}{\checkmark} & \textcolor{customred}{\texttimes} & 0.00 & 0.00 & 0.00 \\
\midrule

\multirow{8}{*}{\makecell[l]{Watching\\TV}} 
& \textcolor{customblue}{\checkmark} & \textcolor{customblue}{\checkmark} & \textcolor{customblue}{\checkmark} & \textcolor{customblue}{\checkmark} & 0.88 & 0.70 & 0.78 \\
& \textcolor{customblue}{\checkmark} & \textcolor{customblue}{\checkmark} & \textcolor{customblue}{\checkmark} & \textcolor{customred}{\texttimes} & 0.73 & 0.80 & 0.76 \\
& \textcolor{customblue}{\checkmark} & \textcolor{customred}{\texttimes} & \textcolor{customred}{\texttimes} & \textcolor{customred}{\texttimes} & 0.50 & 0.80 & 0.62 \\
& \textcolor{customred}{\texttimes} & \textcolor{customblue}{\checkmark} & \textcolor{customred}{\texttimes} & \textcolor{customred}{\texttimes} & 0.19 & 0.70 & 0.30 \\
& \textcolor{customred}{\texttimes} & \textcolor{customred}{\texttimes} & \textcolor{customblue}{\checkmark} & \textcolor{customred}{\texttimes} & 0.00 & 0.00 & 0.00 \\
\midrule

\multirow{8}{*}{\makecell[l]{Workout/\\Weightlifting}} 
& \textcolor{customblue}{\checkmark} & \textcolor{customblue}{\checkmark} & \textcolor{customblue}{\checkmark} & \textcolor{customblue}{\checkmark} & 0.48 & 1.00 & 0.65 \\
& \textcolor{customblue}{\checkmark} & \textcolor{customblue}{\checkmark} & \textcolor{customblue}{\checkmark} & \textcolor{customred}{\texttimes} & 0.30 & 0.60 & 0.40 \\
& \textcolor{customblue}{\checkmark} & \textcolor{customred}{\texttimes} & \textcolor{customred}{\texttimes} & \textcolor{customred}{\texttimes} & 0.18 & 0.30 & 0.22 \\
& \textcolor{customred}{\texttimes} & \textcolor{customblue}{\checkmark} & \textcolor{customred}{\texttimes} & \textcolor{customred}{\texttimes} & 1.00 & 0.10 & 0.18 \\
& \textcolor{customred}{\texttimes} & \textcolor{customred}{\texttimes} & \textcolor{customblue}{\checkmark} & \textcolor{customred}{\texttimes} & 0.08 & 0.70 & 0.14 \\

\bottomrule
\end{tabular}
\end{table}

%%%%%%%%%%%%%%%%%%%%%%%%%%%%%%%%%%%%%%%%%%%%%
%%%%%      QWEN PER-ACTIVITY TABLE      %%%%%
%%%%%%%%%%%%%%%%%%%%%%%%%%%%%%%%%%%%%%%%%%%%%

%\renewcommand{\arraystretch}{0.6}
\begin{table}[H]
\centering
\caption{Closed-set one-shot per-activity performance with \texttt{Qwen-32B}.}
\label{tab:per_task_ablation_qwen}
\setlength{\tabcolsep}{5pt}
\footnotesize
\begin{tabular}{lcccccccc}
\toprule
\multirow{3}{*}{\makecell{\textbf{High-level}\\\textbf{Activity}}} & \multicolumn{4}{c}{\textbf{Modalities}} & \multirow{3}{*}{\textbf{Precision}} & \multirow{3}{*}{\textbf{Recall}}& \multirow{3}{*}{\textbf{F1}} \\
\cmidrule(lr){2-5}
& \makecell{\textbf{Audio}\\\textbf{Caption}} & \makecell{\textbf{Audio}\\\textbf{Labels}} & \makecell{\textbf{IMU Activity}\\\textbf{Pred.}} & \makecell{\textbf{Extra}\\\textbf{Context}} &  \\
\midrule

\multirow{8}{*}{\makecell[l]{Vacuum\\Cleaning}} 
& \textcolor{customblue}{\checkmark} & \textcolor{customblue}{\checkmark} & \textcolor{customblue}{\checkmark} & \textcolor{customblue}{\checkmark} & 0.21 & 0.50 & 0.29 \\
& \textcolor{customblue}{\checkmark} & \textcolor{customblue}{\checkmark} & \textcolor{customblue}{\checkmark} & \textcolor{customred}{\texttimes} & 0.21 & 0.50 & 0.29 \\

& \textcolor{customblue}{\checkmark} & \textcolor{customred}{\texttimes} & \textcolor{customred}{\texttimes} & \textcolor{customred}{\texttimes} & 0.20 & 0.40 & 0.27 \\
& \textcolor{customred}{\texttimes} & \textcolor{customblue}{\checkmark} & \textcolor{customred}{\texttimes} & \textcolor{customred}{\texttimes} & 0.33 & 0.30 & 0.32 \\
& \textcolor{customred}{\texttimes} & \textcolor{customred}{\texttimes} & \textcolor{customblue}{\checkmark} & \textcolor{customred}{\texttimes} & 0.12 & 1.00 & 0.21 \\
\midrule

\multirow{8}{*}{Cooking} 
& \textcolor{customblue}{\checkmark} & \textcolor{customblue}{\checkmark} & \textcolor{customblue}{\checkmark} & \textcolor{customblue}{\checkmark} & 0.58 & 0.70 & 0.64 \\
& \textcolor{customblue}{\checkmark} & \textcolor{customblue}{\checkmark} & \textcolor{customblue}{\checkmark} & \textcolor{customred}{\texttimes} & 0.83 & 0.50 & 0.62 \\
% & \textcolor{customblue}{\checkmark} & \textcolor{customblue}{\checkmark} & \textcolor{customred}{\texttimes} & \textcolor{customred}{\texttimes} & 0.60 & 0.60 & 0.60 \\
% & \textcolor{customblue}{\checkmark} & \textcolor{customred}{\texttimes} & \textcolor{customblue}{\checkmark} & \textcolor{customred}{\texttimes} & 0.70 & 0.70 & 0.70 \\
% & \textcolor{customred}{\texttimes} & \textcolor{customblue}{\checkmark} & \textcolor{customblue}{\checkmark} & \textcolor{customred}{\texttimes} & 0.35 & 0.60 & 0.44 \\
& \textcolor{customblue}{\checkmark} & \textcolor{customred}{\texttimes} & \textcolor{customred}{\texttimes} & \textcolor{customred}{\texttimes} & 0.70 & 0.70 & 0.70 \\
& \textcolor{customred}{\texttimes} & \textcolor{customblue}{\checkmark} & \textcolor{customred}{\texttimes} & \textcolor{customred}{\texttimes} & 0.47 & 0.70 & 0.56 \\
& \textcolor{customred}{\texttimes} & \textcolor{customred}{\texttimes} & \textcolor{customblue}{\checkmark} & \textcolor{customred}{\texttimes} & 0.00 & 0.00 & 0.00 \\
\midrule

\multirow{8}{*}{\makecell[l]{Doing\\laundry}} 
& \textcolor{customblue}{\checkmark} & \textcolor{customblue}{\checkmark} & \textcolor{customblue}{\checkmark} & \textcolor{customblue}{\checkmark} & 1.00 & 0.30 & 0.46 \\
& \textcolor{customblue}{\checkmark} & \textcolor{customblue}{\checkmark} & \textcolor{customblue}{\checkmark} & \textcolor{customred}{\texttimes} & 0.20 & 0.10 & 0.13 \\

& \textcolor{customblue}{\checkmark} & \textcolor{customred}{\texttimes} & \textcolor{customred}{\texttimes} & \textcolor{customred}{\texttimes} & 1.00 & 0.20 & 0.33 \\
& \textcolor{customred}{\texttimes} & \textcolor{customblue}{\checkmark} & \textcolor{customred}{\texttimes} & \textcolor{customred}{\texttimes} & 0.00 & 0.00 & 0.00 \\
& \textcolor{customred}{\texttimes} & \textcolor{customred}{\texttimes} & \textcolor{customblue}{\checkmark} & \textcolor{customred}{\texttimes} & 0.00 & 0.00 & 0.00 \\
\midrule

\multirow{8}{*}{Eating} 
& \textcolor{customblue}{\checkmark} & \textcolor{customblue}{\checkmark} & \textcolor{customblue}{\checkmark} & \textcolor{customblue}{\checkmark} & 1.00 & 0.50 & 0.67 \\
& \textcolor{customblue}{\checkmark} & \textcolor{customblue}{\checkmark} & \textcolor{customblue}{\checkmark} & \textcolor{customred}{\texttimes} & 0.71 & 0.50 & 0.59 \\

& \textcolor{customblue}{\checkmark} & \textcolor{customred}{\texttimes} & \textcolor{customred}{\texttimes} & \textcolor{customred}{\texttimes} & 0.75 & 0.30 & 0.43 \\
& \textcolor{customred}{\texttimes} & \textcolor{customblue}{\checkmark} & \textcolor{customred}{\texttimes} & \textcolor{customred}{\texttimes} & 0.50 & 0.70 & 0.58 \\
& \textcolor{customred}{\texttimes} & \textcolor{customred}{\texttimes} & \textcolor{customblue}{\checkmark} & \textcolor{customred}{\texttimes} & 0.00 & 0.00 & 0.00 \\
\midrule

\multirow{8}{*}{\makecell[l]{Playing\\basketball}} 
& \textcolor{customblue}{\checkmark} & \textcolor{customblue}{\checkmark} & \textcolor{customblue}{\checkmark} & \textcolor{customblue}{\checkmark} & 0.80 & 0.80 & 0.80 \\
& \textcolor{customblue}{\checkmark} & \textcolor{customblue}{\checkmark} & \textcolor{customblue}{\checkmark} & \textcolor{customred}{\texttimes} & 1.00 & 0.80 & 0.89 \\
% & \textcolor{customblue}{\checkmark} & \textcolor{customblue}{\checkmark} & \textcolor{customred}{\texttimes} & \textcolor{customred}{\texttimes} & 0.78 & 0.70 & 0.74 \\
% & \textcolor{customblue}{\checkmark} & \textcolor{customred}{\texttimes} & \textcolor{customblue}{\checkmark} & \textcolor{customred}{\texttimes} & 0.83 & 0.50 & 0.62 \\
% & \textcolor{customred}{\texttimes} & \textcolor{customblue}{\checkmark} & \textcolor{customblue}{\checkmark} & \textcolor{customred}{\texttimes} & 0.50 & 0.40 & 0.44 \\
& \textcolor{customblue}{\checkmark} & \textcolor{customred}{\texttimes} & \textcolor{customred}{\texttimes} & \textcolor{customred}{\texttimes} & 0.86 & 0.60 & 0.71 \\
& \textcolor{customred}{\texttimes} & \textcolor{customblue}{\checkmark} & \textcolor{customred}{\texttimes} & \textcolor{customred}{\texttimes} & 0.80 & 0.40 & 0.53 \\
& \textcolor{customred}{\texttimes} & \textcolor{customred}{\texttimes} & \textcolor{customblue}{\checkmark} & \textcolor{customred}{\texttimes} & 0.00 & 0.00 & 0.00 \\
\midrule

\multirow{8}{*}{\makecell[l]{Playing\\soccer}} 
& \textcolor{customblue}{\checkmark} & \textcolor{customblue}{\checkmark} & \textcolor{customblue}{\checkmark} & \textcolor{customblue}{\checkmark} & 0.80 & 0.40 & 0.53 \\
& \textcolor{customblue}{\checkmark} & \textcolor{customblue}{\checkmark} & \textcolor{customblue}{\checkmark} & \textcolor{customred}{\texttimes} & 1.00 & 0.30 & 0.46 \\
% & \textcolor{customblue}{\checkmark} & \textcolor{customblue}{\checkmark} & \textcolor{customred}{\texttimes} & \textcolor{customred}{\texttimes} & 0.75 & 0.30 & 0.43 \\
% & \textcolor{customblue}{\checkmark} & \textcolor{customred}{\texttimes} & \textcolor{customblue}{\checkmark} & \textcolor{customred}{\texttimes} & 0.50 & 0.30 & 0.38 \\
% & \textcolor{customred}{\texttimes} & \textcolor{customblue}{\checkmark} & \textcolor{customblue}{\checkmark} & \textcolor{customred}{\texttimes} & 1.00 & 0.10 & 0.18 \\
& \textcolor{customblue}{\checkmark} & \textcolor{customred}{\texttimes} & \textcolor{customred}{\texttimes} & \textcolor{customred}{\texttimes} & 0.40 & 0.20 & 0.27 \\
& \textcolor{customred}{\texttimes} & \textcolor{customblue}{\checkmark} & \textcolor{customred}{\texttimes} & \textcolor{customred}{\texttimes} & 0.00 & 0.00 & 0.00 \\
& \textcolor{customred}{\texttimes} & \textcolor{customred}{\texttimes} & \textcolor{customblue}{\checkmark} & \textcolor{customred}{\texttimes} & 0.60 & 0.30 & 0.40 \\
\midrule

\multirow{8}{*}{\makecell[l]{Playing\\with pets}} 
& \textcolor{customblue}{\checkmark} & \textcolor{customblue}{\checkmark} & \textcolor{customblue}{\checkmark} & \textcolor{customblue}{\checkmark} & 0.62 & 1.00 & 0.77 \\
& \textcolor{customblue}{\checkmark} & \textcolor{customblue}{\checkmark} & \textcolor{customblue}{\checkmark} & \textcolor{customred}{\texttimes} & 0.59 & 1.00 & 0.74 \\
% & \textcolor{customblue}{\checkmark} & \textcolor{customblue}{\checkmark} & \textcolor{customred}{\texttimes} & \textcolor{customred}{\texttimes} & 0.50 & 0.90 & 0.64 \\
% & \textcolor{customblue}{\checkmark} & \textcolor{customred}{\texttimes} & \textcolor{customblue}{\checkmark} & \textcolor{customred}{\texttimes} & 0.75 & 0.90 & 0.82 \\
% & \textcolor{customred}{\texttimes} & \textcolor{customblue}{\checkmark} & \textcolor{customblue}{\checkmark} & \textcolor{customred}{\texttimes} & 0.39 & 0.70 & 0.50 \\
& \textcolor{customblue}{\checkmark} & \textcolor{customred}{\texttimes} & \textcolor{customred}{\texttimes} & \textcolor{customred}{\texttimes} & 0.77 & 1.00 & 0.87 \\
& \textcolor{customred}{\texttimes} & \textcolor{customblue}{\checkmark} & \textcolor{customred}{\texttimes} & \textcolor{customred}{\texttimes} & 0.33 & 0.60 & 0.43 \\
& \textcolor{customred}{\texttimes} & \textcolor{customred}{\texttimes} & \textcolor{customblue}{\checkmark} & \textcolor{customred}{\texttimes} & 0.00 & 0.00 & 0.00 \\
\midrule

\multirow{8}{*}{\makecell[l]{Reading\\a book}} 
& \textcolor{customblue}{\checkmark} & \textcolor{customblue}{\checkmark} & \textcolor{customblue}{\checkmark} & \textcolor{customblue}{\checkmark} & 0.60 & 0.30 & 0.40 \\
& \textcolor{customblue}{\checkmark} & \textcolor{customblue}{\checkmark} & \textcolor{customblue}{\checkmark} & \textcolor{customred}{\texttimes} & 1.00 & 0.30 & 0.46 \\
% & \textcolor{customblue}{\checkmark} & \textcolor{customblue}{\checkmark} & \textcolor{customred}{\texttimes} & \textcolor{customred}{\texttimes} & 0.67 & 0.40 & 0.50 \\
% & \textcolor{customblue}{\checkmark} & \textcolor{customred}{\texttimes} & \textcolor{customblue}{\checkmark} & \textcolor{customred}{\texttimes} & 1.00 & 0.20 & 0.33 \\
% & \textcolor{customred}{\texttimes} & \textcolor{customblue}{\checkmark} & \textcolor{customblue}{\checkmark} & \textcolor{customred}{\texttimes} & 0.60 & 0.30 & 0.40 \\
& \textcolor{customblue}{\checkmark} & \textcolor{customred}{\texttimes} & \textcolor{customred}{\texttimes} & \textcolor{customred}{\texttimes} & 0.80 & 0.40 & 0.53 \\
& \textcolor{customred}{\texttimes} & \textcolor{customblue}{\checkmark} & \textcolor{customred}{\texttimes} & \textcolor{customred}{\texttimes} & 0.25 & 0.20 & 0.22 \\
& \textcolor{customred}{\texttimes} & \textcolor{customred}{\texttimes} & \textcolor{customblue}{\checkmark} & \textcolor{customred}{\texttimes} & 0.00 & 0.00 & 0.00 \\
\midrule

\multirow{8}{*}{\makecell[l]{Using\\computer}} 
& \textcolor{customblue}{\checkmark} & \textcolor{customblue}{\checkmark} & \textcolor{customblue}{\checkmark} & \textcolor{customblue}{\checkmark} & 0.50 & 0.70 & 0.58 \\
& \textcolor{customblue}{\checkmark} & \textcolor{customblue}{\checkmark} & \textcolor{customblue}{\checkmark} & \textcolor{customred}{\texttimes} & 0.53 & 0.80 & 0.64 \\
% & \textcolor{customblue}{\checkmark} & \textcolor{customblue}{\checkmark} & \textcolor{customred}{\texttimes} & \textcolor{customred}{\texttimes} & 0.40 & 0.60 & 0.48 \\
% & \textcolor{customblue}{\checkmark} & \textcolor{customred}{\texttimes} & \textcolor{customblue}{\checkmark} & \textcolor{customred}{\texttimes} & 0.35 & 0.60 & 0.44 \\
% & \textcolor{customred}{\texttimes} & \textcolor{customblue}{\checkmark} & \textcolor{customblue}{\checkmark} & \textcolor{customred}{\texttimes} & 0.50 & 0.60 & 0.55 \\
& \textcolor{customblue}{\checkmark} & \textcolor{customred}{\texttimes} & \textcolor{customred}{\texttimes} & \textcolor{customred}{\texttimes} & 0.40 & 0.60 & 0.48 \\
& \textcolor{customred}{\texttimes} & \textcolor{customblue}{\checkmark} & \textcolor{customred}{\texttimes} & \textcolor{customred}{\texttimes} & 0.29 & 0.50 & 0.37 \\
& \textcolor{customred}{\texttimes} & \textcolor{customred}{\texttimes} & \textcolor{customblue}{\checkmark} & \textcolor{customred}{\texttimes} & 0.00 & 0.00 & 0.00 \\
\midrule

\multirow{8}{*}{\makecell[l]{Washing\\dishes}} 
& \textcolor{customblue}{\checkmark} & \textcolor{customblue}{\checkmark} & \textcolor{customblue}{\checkmark} & \textcolor{customblue}{\checkmark} & 0.89 & 0.80 & 0.84 \\
& \textcolor{customblue}{\checkmark} & \textcolor{customblue}{\checkmark} & \textcolor{customblue}{\checkmark} & \textcolor{customred}{\texttimes} & 0.67 & 0.80 & 0.73 \\
% & \textcolor{customblue}{\checkmark} & \textcolor{customblue}{\checkmark} & \textcolor{customred}{\texttimes} & \textcolor{customred}{\texttimes} & 0.70 & 0.70 & 0.70 \\
% & \textcolor{customblue}{\checkmark} & \textcolor{customred}{\texttimes} & \textcolor{customblue}{\checkmark} & \textcolor{customred}{\texttimes} & 0.50 & 0.50 & 0.50 \\
% & \textcolor{customred}{\texttimes} & \textcolor{customblue}{\checkmark} & \textcolor{customblue}{\checkmark} & \textcolor{customred}{\texttimes} & 0.64 & 0.70 & 0.67 \\
& \textcolor{customblue}{\checkmark} & \textcolor{customred}{\texttimes} & \textcolor{customred}{\texttimes} & \textcolor{customred}{\texttimes} & 0.88 & 0.70 & 0.78 \\
& \textcolor{customred}{\texttimes} & \textcolor{customblue}{\checkmark} & \textcolor{customred}{\texttimes} & \textcolor{customred}{\texttimes} & 0.82 & 0.90 & 0.86 \\
& \textcolor{customred}{\texttimes} & \textcolor{customred}{\texttimes} & \textcolor{customblue}{\checkmark} & \textcolor{customred}{\texttimes} & 0.00 & 0.00 & 0.00 \\
\midrule

\multirow{8}{*}{\makecell[l]{Watching\\TV}} 
& \textcolor{customblue}{\checkmark} & \textcolor{customblue}{\checkmark} & \textcolor{customblue}{\checkmark} & \textcolor{customblue}{\checkmark} & 0.67 & 0.20 & 0.31 \\
& \textcolor{customblue}{\checkmark} & \textcolor{customblue}{\checkmark} & \textcolor{customblue}{\checkmark} & \textcolor{customred}{\texttimes} & 0.67 & 0.20 & 0.31 \\
% & \textcolor{customblue}{\checkmark} & \textcolor{customblue}{\checkmark} & \textcolor{customred}{\texttimes} & \textcolor{customred}{\texttimes} & 0.57 & 0.40 & 0.47 \\
% & \textcolor{customblue}{\checkmark} & \textcolor{customred}{\texttimes} & \textcolor{customblue}{\checkmark} & \textcolor{customred}{\texttimes} & 0.71 & 0.50 & 0.59 \\
% & \textcolor{customred}{\texttimes} & \textcolor{customblue}{\checkmark} & \textcolor{customblue}{\checkmark} & \textcolor{customred}{\texttimes} & 0.80 & 0.40 & 0.53 \\
& \textcolor{customblue}{\checkmark} & \textcolor{customred}{\texttimes} & \textcolor{customred}{\texttimes} & \textcolor{customred}{\texttimes} & 0.43 & 0.60 & 0.50 \\
& \textcolor{customred}{\texttimes} & \textcolor{customblue}{\checkmark} & \textcolor{customred}{\texttimes} & \textcolor{customred}{\texttimes} & 0.39 & 0.70 & 0.50 \\
& \textcolor{customred}{\texttimes} & \textcolor{customred}{\texttimes} & \textcolor{customblue}{\checkmark} & \textcolor{customred}{\texttimes} & 1.00 & 0.10 & 0.18 \\
\midrule

\multirow{8}{*}{\makecell[l]{Workout/\\Weightlifting}} 
& \textcolor{customblue}{\checkmark} & \textcolor{customblue}{\checkmark} & \textcolor{customblue}{\checkmark} & \textcolor{customblue}{\checkmark} & 0.36 & 0.50 & 0.42 \\
& \textcolor{customblue}{\checkmark} & \textcolor{customblue}{\checkmark} & \textcolor{customblue}{\checkmark} & \textcolor{customred}{\texttimes} & 0.18 & 0.30 & 0.22 \\
% & \textcolor{customblue}{\checkmark} & \textcolor{customblue}{\checkmark} & \textcolor{customred}{\texttimes} & \textcolor{customred}{\texttimes} & 0.12 & 0.10 & 0.11 \\
% & \textcolor{customblue}{\checkmark} & \textcolor{customred}{\texttimes} & \textcolor{customblue}{\checkmark} & \textcolor{customred}{\texttimes} & 0.14 & 0.20 & 0.17 \\
% & \textcolor{customred}{\texttimes} & \textcolor{customblue}{\checkmark} & \textcolor{customblue}{\checkmark} & \textcolor{customred}{\texttimes} & 0.33 & 0.30 & 0.32 \\
& \textcolor{customblue}{\checkmark} & \textcolor{customred}{\texttimes} & \textcolor{customred}{\texttimes} & \textcolor{customred}{\texttimes} & 0.29 & 0.50 & 0.37 \\
& \textcolor{customred}{\texttimes} & \textcolor{customblue}{\checkmark} & \textcolor{customred}{\texttimes} & \textcolor{customred}{\texttimes} & 0.00 & 0.00 & 0.00 \\
& \textcolor{customred}{\texttimes} & \textcolor{customred}{\texttimes} & \textcolor{customblue}{\checkmark} & \textcolor{customred}{\texttimes} & 0.12 & 0.20 & 0.15 \\

\bottomrule
\end{tabular}
\end{table}
\clearpage

\section{LLM Prompts} \label{sec:prompts}
\begin{tcolorbox}[title=Prompt for activity classification via late fusion=false, colback=gray!5!white, colframe=black, width=\textwidth, sharp corners, boxrule=0.5mm]
RESPOND IN ENGLISH ONLY.  KEEP YOUR ANSWER CONCISE.\\
Below are temporally ordered 'audio captions' and 'top-5 audio labels' generated based on the sounds captured from the environment.  Information under every single 'Time step' is based on 2-second audio recordings that follow each other.Audio information is accompanied by 'physical activity' predictions based on data recorded from a 'head-mounted' inertia measurement unit (IMU).  At each 'Time step', we use a machine learning model to predict one of the following six labels: 'walking', 'running', 'standing', 'bend over', 'stand up', 'sit down'.  When a predictions is 'Not available', it simply means that the model used for making that prediction (e.g., IMU data for activity prediction) was not available for that instance.  For instance, if audio captions are 'Not available' for an instance, it means an audio captioning model was not available. It does NOT mean there was no sound.  Or, if IMU model prediction is 'Not available' for an instance, it means an IMU physical activity prediction model was not available. It does NOT mean there was no or little physical activity.  
\\
\\ Temporally ordered audio captions, audio labels, and IMU physical activity tags: \textcolor{blue}{\{Per modality time series results for each time step\}}
\\
\\
IMPORTANT NOTE: The machine learning models that makes the audio and physical activity predictions are not perfect. They can make mistakes.  For instance, the audio model can mistake the sound of wiping the windows for the sound of a DJ spinning the vinyl.  Or, the physical activity prediction model can mistake that someone is running when they are walking fast.  Therefore, do not hyperfocus on specific audio and physical activity tags.  Focus on reasoning about the underlying sounds and physical motions that would lead to model predictions.  Use information from across different timesteps and modalities to form a more robust big picture.  In light of the information provided above, choose the most likely 'high-level activity' the camera wearer might be doing from.  Think step by step and briefly explain your reasoning behind consistently combining the information across time steps.  Finally, after reasoning, respond with the name of the activity at the end.
\\

\textcolor{blue}{\{For closed set evaluation\}}
List of activities to choose from (please choose ONLY ONE and reply with the exact same name as in the list.):
Playing soccer\\
Playing basketball\\
Cooking\\
Cleaning\\
Eating\\
Washing dishes\\
Doing laundry\\
Reading a book\\
Using computer\\
Watching TV\\
Workout/Weightlifting\\
Playing with pets\\
\\

% I want to generate synthetic data to test LLM's ability to evaluate instruction following. There are three types of responses in instruction-following; correct, subtly off-target, totally wrong. Your task involves generating subtly off-target responses that are subtly disregard the instruction. Subtly off-target response should be classified as false according to pre-defined evaluation functions based on evaluation focus provided below.
% For example, if the instruction is ``Make sure to include at least 10 placeholders...'' and kwargs is ``number-placeholders 10'' then subtly off-target response might include only 8 placeholders. Another example, if the instruction is ``Make sure to include keywords: love, hate'' and kwargs is ``keywords:love, hate'', then subtly off-target response may include only one of them.
% \\
% \\
% Make sure the length of your generated subtly off-target response matches the length of the correct response provided below, aiming for a similar level of detail and complexity.
% \tcblower
% \# Instruction: \textcolor{blue}{\{Instruction-and-task input\}}\\
% \# Evaluation focus: \textcolor{blue}{\{Type\}}\\
% \# Details of evaluation criterion: \textcolor{blue}{\{Kwargs\}}\\
% \# Correctly followed response: \textcolor{blue}{\{Correct Response\}}\\
% \# Subtly off-target response: 
\end{tcolorbox}
\begin{tcolorbox}[title=One shot example=false, colback=gray!5!white, colframe=black, width=\textwidth, sharp corners, boxrule=0.5mm]
\scriptsize
I will give one illustrative example with physical activity and audio predictions, and some reasoning based on it to infer the high-level activity.\\
--- Beginning of temporal audio and physical activity predictions for the illustrative example---\\
\textcolor{blue}{\{--- Time step 1 ---\}}
Top-5 audio labels with probabilities:
Breaking: 0.51, Door: 0.15, Slam: 0.10, Speech: 0.03, Thunk: 0.02\\
Audio caption:
Someone is banging pots and pans together \\
IMU model prediction:
stand up\\
\textcolor{blue}{\{--- Time step 2 ---\}}
Top-5 audio labels with probabilities:
Scrape: 0.22, Zipper (clothing): 0.17, Rub: 0.16, Wood: 0.07, Door: 0.03, \\
Audio caption:
Someone is opening and closing a drawer. \\
IMU model prediction:
running\\
\textcolor{blue}{\{--- Time step 3 ---\}}
Top-5 audio labels with probabilities:
Rub: 0.62, Wood: 0.20, Sawing: 0.09, Sanding: 0.03, Chopping (food): 0.01, \\
Audio caption:
Someone is cleaning a surface with a broom. \\
IMU model prediction:
standing\\
\textcolor{blue}{\{--- Time step 4 ---\}}
Top-5 audio labels with probabilities:
Rub: 0.24, Chopping (food): 0.12, Wood: 0.07, Sanding: 0.04, Tearing: 0.04, \\
Audio caption:
Someone is cleaning a surface with a hammer. \\
IMU model prediction:
walking\\
\textcolor{blue}{\{--- Time step 5 ---\}}
Top-5 audio labels with probabilities:
Bathtub (filling or washing): 0.41, Water: 0.14, Water tap, faucet: 0.05, Speech: 0.04, Drip: 0.04, \\
Audio caption:
A person is washing dishes in a sink. \\
IMU model prediction:
walking\\
\textcolor{blue}{\{--- Time step 6 ---\}}
Top-5 audio labels with probabilities:
Rub: 0.25, Gasp: 0.12, Speech: 0.10, Sound effect: 0.04, Snort: 0.03, \\
Audio caption:
Something is rubbing against a surface \\
IMU model prediction:
walking\\
\textcolor{blue}{\{--- Time step 7 ---\}}
Top-5 audio labels with probabilities:
Door: 0.20, Water: 0.11, Slam: 0.09, Inside, small room: 0.08, Speech: 0.07, \\
Audio caption:
Someone is washing dishes in a sink and then moving them around. \\
IMU model prediction:
stand up\\
\textcolor{blue}{\{--- Time step 8 ---\}}
Top-5 audio labels with probabilities:
Liquid: 0.21, Water: 0.19, Drip: 0.17, Inside, small room: 0.08, Speech: 0.05, \\
Audio caption:
Water is splashing and clacking together as someone is washing dishes. \\
IMU model prediction:
stand up\\
\textcolor{blue}{\{--- Time step 9 ---\}}
Top-5 audio labels with probabilities:
Bathtub (filling or washing): 0.13, Drip: 0.12, Water tap, faucet: 0.12, Water: 0.12, Liquid: 0.08, \\
Audio caption:
Water is splashing and splashing \\
IMU model prediction:
stand up\\
--- End of temporal audio and physical activity predictions for the illustrative example---\\
--- Beginning of reasoning example for the illustrative example--\\
To determine the most likely high-level activity, I analyzed the temporal sequence of audio captions, audio labels, and IMU physical activity predictions across the 9 time steps. Each step represents 2 seconds of data, covering about 18 seconds in total. I focused on consistent patterns and underlying activities, considering the imperfection of the models (e.g., audio labels like "cleaning with a hammer" may be errors, and IMU predictions like "running" during drawer use are likely mismatches). Key insights:\\
- \textbf{Audio Captions and Labels Dominance:} From Time Step 5 onward (Steps 5–9), the audio captions explicitly describe dishwashing activities ("washing dishes in a sink," "moving them around," "water splashing and clacking"). This is supported by high-probability audio labels like "Bathtub (filling or washing)" (Step 5), "Liquid" and "Water" (Steps 7–9), which are characteristic of sink-based tasks. Earlier steps (1–4) include kitchen-like sounds (e.g., "banging pots and pans," "opening and closing a drawer," "cleaning a surface"), but these are precursors or related to the dishwashing theme (e.g., preparing or cleaning up).\\
- \textbf{IMU Physical Activity Integration:} The IMU predictions show frequent "stand up" (Steps 1, 7, 8, 9) and "walking" (Steps 4, 5, 6), which align with the physical motions of dishwashing—such as standing up from bending to retrieve items, walking to fetch or store dishes, or shifting position at the sink. "Standing" (Step 3) also fits. Outliers like "running" (Step 2) are likely model errors given the context (drawer opening doesn't involve running), so they were discounted.\\
- \textbf{Temporal Consistency:} The sequence evolves from general kitchen sounds (Steps 1–4) to sustained dishwashing focus (Steps 5–9), suggesting a single, cohesive activity like washing dishes. For instance:
  - Step 1 (stand up with pot banging) could be rising to start cleaning.
  - Steps 2–3 (drawer, cleaning with broom) might involve gathering tools or pre-cleaning.
  - Step 4's "cleaning with a hammer" is probably a misclassification (e.g., chopping food during prep, but dishwashing dominates later).
  - Steps 5–9 center on water, splashing, and dish movement, with repeated "stand up" indicating motions like lifting dishes or changing posture.\\
- \textbf{Activity Selection:} "Washing dishes" is the clearest match because:
  - It directly explains the water-related sounds and captions from Steps 5–9.
  - The IMU predictions (standing, walking, stand up) are consistent with the intermittent movements in dishwashing (e.g., standing at the sink, walking to put dishes away).
  - Kitchen elements (pots, drawers) from earlier steps are common in dishwashing contexts, as it often follows cooking or involves tidying.\\
Thus, "Washing dishes" best explains the consistent audio evidence and physical motions.\\
user activity: Washing dishes

--- End of reasoning example for the illustrative example\\
Now, please come up with an answer for the original test example I included at the beginning.

\end{tcolorbox}

\section{Selected examples} \label{sec:examples}
% \begin{tcolorbox}[title=Example input and output -- correct prediction (Closed set, one-shot, Gemini); 'Workout/weightlifting'=false, colback=gray!5!white, colframe=black, width=\textwidth, sharp corners, boxrule=0.5mm]

\begin{tcolorbox}[title=Correct prediction; Gemini; closed set ('Workout/weightlifting')=false, colback=gray!5!white, colframe=black, width=\textwidth, sharp corners, boxrule=0.5mm]
\tiny
\textbf{Input}
--- Beginning of temporal audio and physical activity predictions for the example---\\
\textcolor{blue}{\{--- Time step 1 ---\}}\\
Top-5 audio labels with probabilities:
Gasp: 0.56, Snort: 0.15, Speech: 0.04, Breathing: 0.03, Sneeze: 0.02, \\
Audio caption:
A person is walking around and breathing heavily\\ 
IMU model prediction:
stand up\\
\textcolor{blue}{\{--- Time step 2 ---\}}\\
Top-5 audio labels with probabilities:
Chewing, mastication: 0.20, Inside, small room: 0.12, Sound effect: 0.07, Speech: 0.04, Biting: 0.04,\\
Audio caption:
A person is digging a hole in a hard surface. \\
IMU model prediction:
stand up\\
\textcolor{blue}{\{--- Time step 3 ---\}}\\
Top-5 audio labels with probabilities:
Crunch: 0.29, Chewing, mastication: 0.14, Biting: 0.10, Inside, small room: 0.09, Sound effect: 0.05, \\
Audio caption:
A person is digging through a hard surface \\
IMU model prediction:
bend over\\
\textcolor{blue}{\{--- Time step 4 ---\}}\\
Top-5 audio labels with probabilities:
Gasp: 0.50, Snort: 0.10, Speech: 0.08, Sigh: 0.07, Inside, small room: 0.03
Audio caption:
A person is breathing heavily and breathing heavily\\
IMU model prediction:
stand up\\
\textcolor{blue}{\{--- Time step 5 ---\}}\\
Top-5 audio labels with probabilities:
Gasp: 0.70, Snort: 0.21, Speech: 0.02, Sigh: 0.01, Breathing: 0.01, \\
Audio caption:
A person is breathing heavily and breathing heavily. \\
IMU model prediction:
stand up\\
\textcolor{blue}{\{--- Time step 6 ---\}}\\
Top-5 audio labels with probabilities:
Sigh: 0.79, Gasp: 0.13, Snort: 0.02, Breathing: 0.01, Speech: 0.01, \\
Audio caption:
A person is breathing heavily and repeatedly. \\
IMU model prediction:
walking\\
\textcolor{blue}{\{--- Time step 7 ---\}}\\
Top-5 audio labels with probabilities:
Sigh: 0.53, Snort: 0.13, Gasp: 0.09, Breathing: 0.08, Sound effect: 0.03, \\
Audio caption:
A person is walking slowly and carefully across a room. \\
IMU model prediction:
walking\\
\textcolor{blue}{\{--- Time step 8 ---\}}\\
Top-5 audio labels with probabilities:
Snort: 0.44, Gasp: 0.15, Speech: 0.07, Sigh: 0.06, Animal: 0.02, \\
Audio caption:
A person breathing heavily \\
IMU model prediction:
bend over\\
\textcolor{blue}{\{--- Time step 9 ---\}}\\
Top-5 audio labels with probabilities:
Gasp: 0.26, Snort: 0.19, Speech: 0.08, Sneeze: 0.07, Breathing: 0.05, \\
Audio caption:
A man coughs and exhales \\
IMU model prediction:
bend over\\
--- End of temporal audio and physical activity predictions ---\\
\\
\textbf{Output}\\
Step-by-Step Reasoning

To determine the most likely high-level activity, I analyzed the sequence of audio and IMU data. The key is to find a consistent theme across both modalities, discounting potential model errors.

1.  **Audio Analysis**: The most dominant and recurring audio theme is strenuous breathing. Across nearly all time steps, the top audio labels include `Gasp`, `Snort`, and `Sigh`, and the captions consistently describe "breathing heavily." This strongly indicates significant physical exertion. The sounds in steps 2 and 3 (`Chewing`, `Crunch`, "digging") are likely misinterpretations of sounds related to effort, such as grunting, teeth clenching, or the sound of equipment.

2.  **IMU Physical Activity Analysis**: The IMU data shows a clear pattern of dynamic movement: `stand up` (steps 1, 2, 4, 5), `bend over` (steps 3, 8, 9), and `walking` (steps 6, 7). This sequence of bending down, rising, and moving around is not random but suggests a cyclical, purposeful physical task.

3.  **Synthesis and Integration**: Combining the audio and IMU data paints a clear picture. The heavy breathing (audio) is the direct result of the physical movements (IMU). The cycle of bending over, standing up, and walking is characteristic of many exercises. For example, one might bend to lift a weight (deadlift, kettlebell swing), stand up during the lift, and then walk or pace between sets.

4.  **Activity Selection**:
    *   Activities like `Cooking` or `Cleaning` would not typically produce such consistent and intense heavy breathing.
    *   `Playing soccer` or `basketball` would likely involve more `running`.
    *   `Workout/Weightlifting` perfectly aligns with both the strenuous breathing sounds and the specific physical motions of bending, standing, and pacing recorded by the IMU.

Therefore, the combination of intense respiratory sounds and a repeated pattern of bending and standing strongly suggests a workout session.\\

\textit{user activity: }Workout/Weightlifting
\end{tcolorbox}
% \begin{tcolorbox}[title=Example input and output -- incorrect prediction (Closed-set, one-shot, Gemini) 'Workout/weightlifting'=false, colback=gray!5!white, colframe=black, width=\textwidth, sharp corners, boxrule=0.5mm]
\begin{tcolorbox}[title=Incorrect prediction; Gemini; closed set ('Workout/weightlifting')=false, colback=gray!5!white, colframe=black, width=\textwidth, sharp corners, boxrule=0.5mm]
\tiny
\textbf{Input}
--- Beginning of temporal audio and physical activity predictions for the example---\\
\textcolor{blue}{\{--- Time step 1 ---\}}

Top-5 audio labels with probabilities:
Animal: 0.13, Speech: 0.13, Inside, small room: 0.07, Walk, footsteps: 0.05, Clip-clop: 0.03, 

Audio caption:
A horse walks around 

IMU model prediction:
stand up

\textcolor{blue}{\{--- Time step 2 ---\}}

Top-5 audio labels with probabilities:
Tearing: 0.30, Zipper (clothing): 0.19, Inside, small room: 0.13, Speech: 0.05, Scrape: 0.02, 

Audio caption:
A person rubs a surface and then rubs a surface again. 

IMU model prediction:
stand up

\textcolor{blue}{\{--- Time step 3 ---\}}

Top-5 audio labels with probabilities:
Thunk: 0.15, Speech: 0.13, Inside, small room: 0.11, Breaking: 0.09, Door: 0.07, 

Audio caption:
A person is moving objects around. 

IMU model prediction:
stand up

\textcolor{blue}{\{--- Time step 4 ---\}}

Top-5 audio labels with probabilities:
Inside, small room: 0.19, Scrape: 0.10, Speech: 0.09, Gasp: 0.08, Snort: 0.05, 

Audio caption:
A person breathes and breathes 

IMU model prediction:
stand up

\textcolor{blue}{\{--- Time step 5 ---\}}

Top-5 audio labels with probabilities:
Inside, small room: 0.16, Biting: 0.11, Scrape: 0.07, Speech: 0.05, Door: 0.04, 

Audio caption:
A person is shuffling a piece of wood. 

IMU model prediction:
bend over

\textcolor{blue}{\{--- Time step 6 ---\}}

Top-5 audio labels with probabilities:
Snort: 0.51, Gasp: 0.26, Breathing: 0.03, Speech: 0.02, Sigh: 0.01, 

Audio caption:
A person rubs and breathes 

IMU model prediction:
bend over

\textcolor{blue}{\{--- Time step 7 ---\}}

Top-5 audio labels with probabilities:
Animal: 0.10, Sheep: 0.07, Livestock, farm animals, working animals: 0.06, Snort: 0.06, Speech: 0.05, 

Audio caption:
A person walks on a floor and a door opens. 

IMU model prediction:
bend over

\textcolor{blue}{\{--- Time step 8 ---\}}

Top-5 audio labels with probabilities:
Snort: 0.35, Gasp: 0.09, Speech: 0.06, Animal: 0.05, Inside, small room: 0.03, 

Audio caption:
A person walks slowly and then stops. 

IMU model prediction:
bend over

\textcolor{blue}{\{--- Time step 9 ---\}}

Top-5 audio labels with probabilities:
Snort: 0.34, Gasp: 0.14, Speech: 0.11, Animal: 0.04, Sigh: 0.02, 

Audio caption:
A person is walking and breathing heavily and then opens a door. 

IMU model prediction:
bend over

--- End of temporal audio and physical activity predictions ---\\
\\
\textbf{Output}\\
Step-by-Step Reasoning
To determine the most likely high-level activity, I analyzed the temporal sequence of audio and IMU data. I focused on identifying consistent patterns across both modalities while accounting for potential model inaccuracies.

1.  **Dominant Audio Cues:** Across multiple time steps, the audio data strongly indicates the presence of an animal. Key labels include "Animal" (Steps 1, 7, 8, 9), "Clip-clop" (Step 1), "Livestock" (Step 7), and most notably, "Snort" which appears with high probability in the second half of the sequence (Steps 6, 7, 8, 9). While the initial caption "A horse walks around" is likely a misinterpretation for an indoor setting, the persistence of animal-related sounds is a crucial clue. Sounds like "rubbing," "shuffling," and heavy "breathing" or "gasping" accompany these animal sounds, suggesting a close interaction.

2.  **Consistent Physical Motion:** The IMU data shows a clear pattern of movement. The person first performs a "stand up" motion (Steps 1-4) and then transitions into a sustained "bend over" posture for the remainder of the sequence (Steps 5-9). This physical sequence of standing and then bending over for a prolonged period aligns well with interacting with something at a low level, such as on the floor.

3.  **Synthesizing Audio and IMU Data:** The combination of sustained bending over and the simultaneous sounds of an animal (snorting, gasping) and interaction (rubbing, shuffling) creates a coherent picture. The person likely stands up, approaches an animal, and then bends over to engage with it. The rubbing sounds could be petting, and the shuffling and breathing/snorting could be from active play.

4.  **Activity Selection:** Comparing this synthesized scenario to the provided list, "Playing with pets" is the most logical fit. It explains the animal sounds, the interaction noises (rubbing), and the physical posture (bending over to be at the pet's level). Other activities like `Cleaning` or `Workout` might involve bending, but they do not account for the prominent and consistent animal-related audio cues.

\textit{user activity:} Playing with pets
\end{tcolorbox}
\begin{tcolorbox}[title=Correct prediction; Gemini; open ended ('Cooking')=false, colback=gray!5!white, colframe=black, width=\textwidth, sharp corners, boxrule=0.5mm]
\tiny
\textbf{Input}
--- Beginning of temporal audio and physical activity predictions for the example---\\
\textcolor{blue}{\{--- Time step 1 ---\}}

Top-5 audio labels with probabilities:
Water: 0.06, Sliding door: 0.06, Door: 0.06, Water tap, faucet: 0.04, Fill (with liquid): 0.04, 

Audio caption:
Food is frying and sizzling 

IMU model prediction:
walking

\textcolor{blue}{\{--- Time step 2 ---\}}

Top-5 audio labels with probabilities:
Speech: 0.07, Door: 0.05, Water: 0.03, Liquid: 0.03, Slam: 0.03, 

Audio caption:
A person is cooking food in a pan and then flipping it over. 

IMU model prediction:
walking

\textcolor{blue}{\{--- Time step 3 ---\}}

Top-5 audio labels with probabilities:
Speech: 0.39, Water: 0.04, Throat clearing: 0.04, Animal: 0.03, Music: 0.03, 

Audio caption:
A steady stream of water is splashing in a frying pan. 

IMU model prediction:
stand up

\textcolor{blue}{\{--- Time step 4 ---\}}

Top-5 audio labels with probabilities:
Squeal: 0.10, Vehicle: 0.07, Water: 0.07, Speech: 0.03, Mechanisms: 0.03, 

Audio caption:
Water is flowing in a faucet and splashing in the background. 

IMU model prediction:
stand up

\textcolor{blue}{\{--- Time step 5 ---\}}

Top-5 audio labels with probabilities:
Water: 0.11, Gush: 0.07, Speech: 0.06, Sizzle: 0.05, Water tap, faucet: 0.04, 

Audio caption:
A frying pan is frying in a pan. 

IMU model prediction:
stand up

\textcolor{blue}{\{--- Time step 6 ---\}}

Top-5 audio labels with probabilities:
Speech: 0.09, Gush: 0.07, Water tap, faucet: 0.06, Water: 0.06, Sink (filling or washing): 0.06, 

Audio caption:
A person is frying food in a pan. 

IMU model prediction:
bend over

\textcolor{blue}{\{--- Time step 7 ---\}}

Top-5 audio labels with probabilities:
Door: 0.13, Slam: 0.07, Dishes, pots, and pans: 0.07, Whack, thwack: 0.05, Speech: 0.05, 

Audio caption:
Something is fried and something is tapped 

IMU model prediction:
walking

\textcolor{blue}{\{--- Time step 8 ---\}}

Top-5 audio labels with probabilities:
Dishes, pots, and pans: 0.18, Door: 0.07, Inside, small room: 0.05, Cutlery, silverware: 0.05, Creak: 0.04, 

Audio caption:
Someone is pounding on a hard surface. 

IMU model prediction:
bend over

\textcolor{blue}{\{--- Time step 9 ---\}}

Top-5 audio labels with probabilities:
Slap, smack: 0.18, Finger snapping: 0.13, Speech: 0.06, Ping: 0.06, Door: 0.02, 

Audio caption:
Someone is tapping a hard object against a hard surface. 

IMU model prediction:
bend over

--- End of temporal audio and physical activity predictions ---\\
\\
\textbf{Output}\\
Step-by-Step Reasoning

To determine the most likely high-level activity, I analyzed the sequence of audio and physical activity data. The analysis focuses on identifying consistent patterns that link the sounds and movements into a coherent event.

1.  **Dominant Audio Theme:** The audio data consistently points to a kitchen environment and the act of cooking. Audio captions repeatedly mention "frying and sizzling" (Step 1), "cooking food in a pan" (Steps 2, 6), and "frying pan" (Steps 3, 5). This is strongly supported by audio labels like "Sizzle" (Step 5), "Dishes, pots, and pans" (Steps 7, 8), and "Water tap, faucet" (Steps 1, 5, 6), which are all characteristic of food preparation.

2.  **Integration of Physical Activity:** The IMU predictions align well with the motions involved in cooking.
    *   `walking` (Steps 1, 2, 7) corresponds to moving around a kitchen, for instance, between the stove, counter, and refrigerator.
    *   `stand up` (Steps 3, 4, 5) is consistent with standing at a stove or counter to monitor or stir food.
    *   `bend over` (Steps 6, 8, 9) fits with actions like getting items from a low cabinet, checking an oven, or leaning over a cutting board. The combination of "pounding" and "tapping" sounds with `bend over` (Steps 8, 9) strongly suggests preparing ingredients on a countertop.

3.  **Temporal Consistency:** The sequence of events tells a logical story of preparing a meal. The activity starts with frying (Steps 1-2), involves adding water or using a faucet (Steps 3-4), and continues with frying (Steps 5-6). The final steps include sounds of utensils and dishes ("tapping," "pounding," "Dishes, pots, and pans"), which could be part of the final preparation or plating process.

4.  **Activity Selection:** The combined evidence from both audio and motion sensors overwhelmingly supports "Cooking." The primary sounds are of frying and food preparation, not cleaning or eating. The physical movements are all typical for a person actively engaged in preparing a meal in a kitchen.

\textit{user activity}: Cooking
\end{tcolorbox}
\begin{tcolorbox}[title=Mostly correct prediction; Gemini; open ended ('Cooking')=false, colback=gray!5!white, colframe=black, width=\textwidth, sharp corners, boxrule=0.5mm]
\tiny
\textbf{Input}
--- Beginning of temporal audio and physical activity predictions for the example---\\
\textcolor{blue}{\{--- Time step 1 ---\}}

Top-5 audio labels with probabilities:
Vehicle: 0.19, Car: 0.08, Speech: 0.05, Water: 0.03, Medium engine (mid frequency): 0.03, 

Audio caption:
A steady stream of sizzling sizzles in the background. 

IMU model prediction:
walking

\textcolor{blue}{\{--- Time step 2 ---\}}

Top-5 audio labels with probabilities:
Sliding door: 0.21, Door: 0.17, Vehicle: 0.09, Slam: 0.07, Speech: 0.07, 

Audio caption:
A person is frying food in a pan and then sizzling it. 

IMU model prediction:
walking

\textcolor{blue}{\{--- Time step 3 ---\}}

Top-5 audio labels with probabilities:
Speech: 0.40, Door: 0.22, Sliding door: 0.06, Vehicle: 0.05, Cupboard open or close: 0.05, 

Audio caption:
A woman is running a heavy machine and then a woman is speaking. 

IMU model prediction:
stand up

\textcolor{blue}{\{--- Time step 4 ---\}}

Top-5 audio labels with probabilities:
Vehicle: 0.09, Car: 0.07, Speech: 0.05, White noise: 0.05, Water: 0.04, 

Audio caption:
A machine is running in the background. 

IMU model prediction:
stand up

\textcolor{blue}{\{--- Time step 5 ---\}}

Top-5 audio labels with probabilities:
Door: 0.20, Car: 0.16, Vehicle: 0.15, Sliding door: 0.13, Speech: 0.04, 

Audio caption:
A motor is running and a motor is moving 

IMU model prediction:
stand up

\textcolor{blue}{\{--- Time step 6 ---\}}

Top-5 audio labels with probabilities:
Vehicle: 0.22, Sliding door: 0.06, Speech: 0.05, Door: 0.04, Car: 0.04, 

Audio caption:
A car engine is running and rattling. 

IMU model prediction:
stand up

\textcolor{blue}{\{--- Time step 7 ---\}}

Top-5 audio labels with probabilities:
Speech: 0.17, Vehicle: 0.10, Water: 0.08, Slosh: 0.08, Music: 0.07, 

Audio caption:
A machine is running and a car is driving. 

IMU model prediction:
walking

\textcolor{blue}{\{--- Time step 8 ---\}}

Top-5 audio labels with probabilities:
Frying (food): 0.32, Sizzle: 0.25, Stir: 0.21, Vehicle: 0.05, Speech: 0.02, 

Audio caption:
A person is frying food in a pan and then sizzling it. 

IMU model prediction:
walking

\textcolor{blue}{\{--- Time step 9 ---\}}

Top-5 audio labels with probabilities:
Breaking: 0.16, Vehicle: 0.09, Scrape: 0.08, Glass: 0.06, Chink, clink: 0.06, 

Audio caption:
A metal object is being clattered against a metal surface. 

IMU model prediction:
walking

--- End of temporal audio and physical activity predictions ---\\
\\
\textbf{Output}\\
Step-by-Step Reasoning

To determine the most likely high-level activity, I analyzed the consistent patterns across the audio and physical activity data.

1.  **Dominant Audio Themes:** Two primary soundscapes emerge from the data. The first is clearly related to cooking, with highly specific labels and captions like "Frying (food)," "Sizzle," and "Stir" appearing multiple times (Steps 1, 2, 8). The second is a persistent background noise identified as a vehicle, with labels like "Vehicle," "Car," and captions mentioning a "motor" or "engine" in nearly every time step (1, 3, 4, 5, 6, 7).

2.  **IMU Physical Activity Integration:** The physical activity consists of `walking` (Steps 1, 2, 7, 8, 9) and `stand up` (Steps 3, 4, 5, 6). This pattern of intermittent walking and shifting posture is consistent with moving around a limited workspace, such as a kitchen or a stall, rather than continuous walking or standing still.

3.  **Temporal Consistency and Synthesis:** The cooking and vehicle sounds occur simultaneously throughout the sequence. For example, Step 2 mentions frying food while vehicle sounds are also detected. Step 8 again highlights frying and sizzling, while the surrounding steps are dominated by vehicle/engine sounds. This co-occurrence suggests the two activities are happening in the same location. The most logical explanation that combines active cooking with the constant sound of a running engine is that the cooking is taking place within or adjacent to a vehicle designed for this purpose. The sounds of a "Sliding door" (Steps 2, 3, 5, 6) could plausibly be the service window of such a vehicle.

4.  **Activity Selection:** While "cooking outdoors" is a possibility, "Cooking in a food truck" provides a more precise explanation for the persistent and prominent "engine" and "motor" sounds described in the captions. This activity perfectly reconciles the specific cooking actions with the constant vehicle noise and the physical movements of working in a confined space.

\textit{user activity}: Cooking in a food truck
\end{tcolorbox}
\begin{tcolorbox}[title=Incorrect prediction; Gemini; open ended ('Playing basketball')=false, colback=gray!5!white, colframe=black, width=\textwidth, sharp corners, boxrule=0.5mm]
\tiny
\textbf{Input}
--- Beginning of temporal audio and physical activity predictions for the example---\\
\textcolor{blue}{\{--- Time step 1 ---\}}

Top-5 audio labels with probabilities:
Speech: 0.12, Gasp: 0.10, Animal: 0.07, Chop: 0.07, Snort: 0.04, 

Audio caption:
Someone is walking on a hard surface. 

IMU model prediction:
stand up

\textcolor{blue}{\{--- Time step 2 ---\}}

Top-5 audio labels with probabilities:
Speech: 0.11, Vehicle: 0.07, Car: 0.06, Door: 0.05, Power windows, electric windows: 0.04, 

Audio caption:
A person is running and walking in a gravel pit. 

IMU model prediction:
running

\textcolor{blue}{\{--- Time step 3 ---\}}

Top-5 audio labels with probabilities:
Slam: 0.25, Speech: 0.06, Basketball bounce: 0.05, Whack, thwack: 0.04, Ping: 0.03, 

Audio caption:
A ball is bouncing and a person is grunting. 

IMU model prediction:
bend over

\textcolor{blue}{\{--- Time step 4 ---\}}

Top-5 audio labels with probabilities:
Speech: 0.14, Door: 0.10, Slam: 0.07, Gasp: 0.04, Chop: 0.04, 

Audio caption:
A rhythmic rhythmic beat is occurring in the background. 

IMU model prediction:
stand up

\textcolor{blue}{\{--- Time step 5 ---\}}

Top-5 audio labels with probabilities:
Chop: 0.65, Speech: 0.03, Whack, thwack: 0.02, Wood: 0.02, Tap: 0.01, 

Audio caption:
A person is walking while a stream of water is flowing. 

IMU model prediction:
stand up

\textcolor{blue}{\{--- Time step 6 ---\}}

Top-5 audio labels with probabilities:
Drum: 0.18, Percussion: 0.13, Musical instrument: 0.13, Music: 0.13, Slam: 0.04, 

Audio caption:
A man is walking and a loud banging sound 

IMU model prediction:
stand up

\textcolor{blue}{\{--- Time step 7 ---\}}

Top-5 audio labels with probabilities:
Door: 0.12, Music: 0.12, Sliding door: 0.08, Slam: 0.07, Speech: 0.06, 

Audio caption:
A ball is being bounced around in a large room. 

IMU model prediction:
stand up

\textcolor{blue}{\{--- Time step 8 ---\}}

Top-5 audio labels with probabilities:
Speech: 0.14, Animal: 0.09, Basketball bounce: 0.04, Tap: 0.03, Slap, smack: 0.03, 

Audio caption:
A ball bouncing in the background 

IMU model prediction:
stand up

\textcolor{blue}{\{--- Time step 9 ---\}}

Top-5 audio labels with probabilities:
Speech: 0.23, Animal: 0.16, Basketball bounce: 0.11, Clip-clop: 0.06, Horse: 0.04, 

Audio caption:
A tennis ball is bouncing back and forth in the background. 

IMU model prediction:
bend over

--- End of temporal audio and physical activity predictions ---\\
\\
\textbf{Output}\\
Step-by-Step Reasoning

To determine the most likely high-level activity, I analyzed the temporal sequence of audio and IMU data. The analysis focuses on identifying consistent patterns across modalities and time steps.

1.  **Dominant Audio Theme: Ball Activity:** A recurring theme across multiple time steps is the sound of a ball. This is explicitly mentioned in the audio captions for steps 3 ("A ball is bouncing"), 7 ("A ball is being bounced around"), 8 ("A ball bouncing"), and 9 ("A tennis ball is bouncing back and forth"). This is further supported by the high-probability audio label "Basketball bounce" appearing in steps 3, 8, and 9.

2.  **Dominant Audio Theme: Impact Sounds:** Several time steps feature sharp, percussive sounds consistent with a sport. These include labels like "Slam" (steps 3, 4, 6), "Whack, thwack" (step 3), and "Chop" (step 5, with a high probability of 0.65). These sounds are likely the impact of a ball being hit, either with a racket or against a surface.

3.  **IMU Physical Activity Integration:** The IMU predictions show a highly active user, inconsistent with a sedentary activity. The sequence includes `running` (step 2), `bend over` (steps 3 and 9), and frequent `stand up` motions. This combination of running, bending, and recovering posture strongly suggests athletic movements, such as chasing a ball, getting into a low stance to hit it, and then standing up. The "grunt" mentioned in the caption for step 3 also indicates physical exertion.

4.  **Synthesizing the Evidence:** The combination of continuous ball bouncing sounds, sharp impact noises ("whack," "chop," "slam"), and dynamic physical movements (running, bending) strongly points to a ball sport. The caption in step 9, "A tennis ball is bouncing back and forth," provides the most specific clue. The "back and forth" motion is characteristic of a racket sport like tennis, rather than dribbling in basketball. The "Whack" and "Chop" sounds are also more indicative of a racket striking a ball. The movements of running and bending are fundamental to playing tennis.

Therefore, the evidence consistently points to the user playing a racket sport, with tennis being the most likely candidate.

\textit{user activity}: Playing tennis
\end{tcolorbox}

\newpage

{
\renewcommand{\arraystretch}{1.2}
\begin{longtable}{llr}
\caption{Ego4D HLA Metadata} \label{tab:ego4d_metadata} \\
\toprule
\textbf{Activity} & \textbf{Video ID} & \textbf{Start Time (s)} \\
\midrule
\endfirsthead
\toprule
\textbf{Activity} & \textbf{Video ID} & \textbf{Start Time (s)} \\
\midrule
\endhead
\bottomrule
\endfoot
reading & ed31f134-b3f0-4f50-b8ae-d956b2f9ccfd & 559.49 \\
reading & 4792d019-ca0c-48fe-b8e6-12e325e8d8ee & 15.07 \\
reading & 4792d019-ca0c-48fe-b8e6-12e325e8d8ee & 80.07 \\
reading & 4792d019-ca0c-48fe-b8e6-12e325e8d8ee & 175.07 \\
reading & f9fa7dac-680a-4301-b783-470aef6f2639 & 1069.98 \\
reading & a5739422-9281-49d2-b15f-2a275cd478d3 & 544.98 \\
reading & a5739422-9281-49d2-b15f-2a275cd478d3 & 684.98 \\
reading & e05c092d-8fd0-467d-a0d5-9e13bfc75b51 & 1679.92 \\
reading & e05c092d-8fd0-467d-a0d5-9e13bfc75b51 & 1754.92 \\
reading & 94e8fb8f-ecf5-409c-940b-a5c966ea2837 & 1425.45 \\
cooking & 59867cc2-3e1f-4dea-b4d7-dbaf55982e72 & 3042.39 \\
cooking & 024713b7-b198-4502-a114-02ca0485353b & 1000.21 \\
cooking & 024713b7-b198-4502-a114-02ca0485353b & 716.26 \\
cooking & 024713b7-b198-4502-a114-02ca0485353b & 819.26 \\
cooking & a16dde53-8a4d-4246-aab4-9759f0af81f0 & 1112.46 \\
cooking & a16dde53-8a4d-4246-aab4-9759f0af81f0 & 1352.46 \\
cooking & 59867cc2-3e1f-4dea-b4d7-dbaf55982e72 & 1135.91 \\
cooking & a16dde53-8a4d-4246-aab4-9759f0af81f0 & 1804.99 \\
cooking & 59867cc2-3e1f-4dea-b4d7-dbaf55982e72 & 3297.39 \\
cooking & a16dde53-8a4d-4246-aab4-9759f0af81f0 & 977.12 \\
laundry & e5bbd2ea-2ffd-4d6a-a187-060e664e30c6 & 0.07 \\
laundry & 6fbe37c5-7321-4223-a26b-78fef89a9c46 & 8.07 \\
laundry & 0ecba8a8-b84b-4020-9487-600f67f4ef5f & 40.07 \\
laundry & 0c892917-34b9-4344-a8cf-d04ff885dc86 & 90.08 \\
laundry & 5059d234-992a-48e6-a9e5-c5777943ecfa & 5.07 \\
laundry & 5afe1e42-d44f-4c1d-bfd1-84de861b4492 & 45.07 \\
laundry & cbe34c13-e91f-4d2f-a902-c011503b6a8f & 0.07 \\
laundry & 01111831-9107-43c4-bf0e-6b26e9b32a2b & 2029.89 \\
laundry & 9feaaf23-d622-479e-9c40-c396e86b4d18 & 88.07 \\
laundry & f442873e-6f2b-493f-a6db-718cd20a732a & 60.02 \\
pets & d45beaf4-ca3c-4a0f-9cee-548d93ae4c71 & 527.36 \\
pets & 7d3d85d3-88e4-4bb9-9175-e37107100f2e & 3.07 \\
pets & 7d3d85d3-88e4-4bb9-9175-e37107100f2e & 75.07 \\
pets & 9d3e9c1a-457e-4125-a847-a535a4c8311f & 200.08 \\
pets & 0296707b-0b42-45dd-98dc-ca8ce0a26a50 & 10.07 \\
pets & 61a437a7-f30e-427f-a5a9-b6d41a7ad499 & 90.07 \\
pets & 35b1a8c7-80fd-402f-b240-768926b7bcee & 80.08 \\
pets & 35b1a8c7-80fd-402f-b240-768926b7bcee & 263.08 \\
pets & b0d3a6de-6520-414d-8161-3ae0ea34feb3 & 0.07 \\
pets & e9725499-415a-490c-a1c7-6089030c958a & 1362.98 \\
soccer & 3ab744ba-9969-452c-a825-38affebd4139 & 559.03 \\
soccer & 3ab744ba-9969-452c-a825-38affebd4139 & 659.03 \\
soccer & 3ab744ba-9969-452c-a825-38affebd4139 & 1394.99 \\
soccer & 3ab744ba-9969-452c-a825-38affebd4139 & 1110.03 \\
soccer & 3ab744ba-9969-452c-a825-38affebd4139 & 1165.03 \\
soccer & 3ab744ba-9969-452c-a825-38affebd4139 & 2170.02 \\
soccer & 3ab744ba-9969-452c-a825-38affebd4139 & 2140.02 \\
soccer & 3ab744ba-9969-452c-a825-38affebd4139 & 816.23 \\
soccer & 3ab744ba-9969-452c-a825-38affebd4139 & 916.23 \\
soccer & 3ab744ba-9969-452c-a825-38affebd4139 & 1046.23 \\
basketball & c48a70f7-44a3-44aa-ac14-baf35e696e5c & 1006.92 \\
basketball & c48a70f7-44a3-44aa-ac14-baf35e696e5c & 1066.92 \\
basketball & c48a70f7-44a3-44aa-ac14-baf35e696e5c & 581.86 \\
basketball & c48a70f7-44a3-44aa-ac14-baf35e696e5c & 686.86 \\
basketball & c48a70f7-44a3-44aa-ac14-baf35e696e5c & 716.86 \\
basketball & a82af40e-2318-42ea-9167-5cc8f8111de1 & 185.07 \\
basketball & a82af40e-2318-42ea-9167-5cc8f8111de1 & 529.53 \\
basketball & c48a70f7-44a3-44aa-ac14-baf35e696e5c & 1377.83 \\
basketball & a82af40e-2318-42ea-9167-5cc8f8111de1 & 706.61 \\
basketball & a82af40e-2318-42ea-9167-5cc8f8111de1 & 796.61 \\
exercise & 3e287c36-44a8-4871-b090-dbd352392181 & 0.07 \\
exercise & 8327741f-773a-4868-902b-0fe25708e0b9 & 65.02 \\
exercise & 8327741f-773a-4868-902b-0fe25708e0b9 & 180.02 \\
exercise & 56d0a3e8-58c0-4d4d-b1a8-097b8cabcc95 & 876.66 \\
exercise & aa873f4c-8100-46fb-83ee-036f908bfa0e & 1944.74 \\
exercise & 7ede3aa4-b631-4715-863e-83cd4ceab976 & 1381.74 \\
exercise & 3d751dba-7897-4ad2-8baa-2569f32dbe6d & 275.08 \\
exercise & 3d751dba-7897-4ad2-8baa-2569f32dbe6d & 568.02 \\
exercise & 3d751dba-7897-4ad2-8baa-2569f32dbe6d & 1352.20 \\
exercise & 13af8d02-be9d-4d80-9999-aa4f31c455b9 & 1016.58 \\
watching\_tv & 3b021bad-6945-4be2-a7db-44af3db1eae8 & 1669.98 \\
watching\_tv & 7a680a9a-dd7b-467a-a724-a597f1be9f06 & 3019.86 \\
watching\_tv & 7a680a9a-dd7b-467a-a724-a597f1be9f06 & 3069.86 \\
watching\_tv & c8d51f6b-7329-4367-9b24-941fa79794d2 & 2449.92 \\
watching\_tv & cf7c12db-1a9e-46d3-96d6-38174bbe373c & 2149.99 \\
watching\_tv & 0793bbe0-b8d5-4d46-9f02-c71d1bd4fad2 & 1729.39 \\
watching\_tv & 5ed59c41-af9e-4368-a0ce-30d536d34895 & 3363.95 \\
watching\_tv & 1bec800a-c3cf-431f-bf0a-7632ad53bcb7 & 2444.99 \\
watching\_tv & dcf7bdf5-2dd1-4334-b79b-7e1963b2dc3c & 2194.89 \\
watching\_tv & dcf7bdf5-2dd1-4334-b79b-7e1963b2dc3c & 2429.89 \\
dishes & ceba5343-c1d5-4f8a-a6bb-ef39c3923e6e & 273.61 \\
dishes & ceba5343-c1d5-4f8a-a6bb-ef39c3923e6e & 358.61 \\
dishes & b79a27cf-4e81-4ae1-8cf8-029711cb1bb7 & 1301.99 \\
dishes & bd16e1a8-5119-4f24-ac6f-d4a388bc78bc & 311.82 \\
dishes & 10a7fd5a-169d-4b0c-ba8b-825af7bd2117 & 1870.02 \\
dishes & f5c456b2-b998-4f42-82bd-786833fb3891 & 165.07 \\
dishes & 58e39e0e-9bbb-4e7f-98b4-14b2fa325fa3 & 115.07 \\
dishes & b6e2e728-11f9-46c0-9fc8-490ea1a905ae & 2529.92 \\
dishes & 30b9ff64-51c8-4481-ba6a-733ea2060aef & 1099.88 \\
dishes & 1fae6ecb-2ad9-4160-b388-c34e7d018915 & 1159.98 \\
using\_pc & f442873e-6f2b-493f-a6db-718cd20a732a & 1079.97 \\
using\_pc & b8276cd7-33e9-4529-80b4-b301b50e227b & 1988.91 \\
using\_pc & b6e2e728-11f9-46c0-9fc8-490ea1a905ae & 2299.83 \\
using\_pc & 876dad36-879c-42eb-977d-b500f7c141d3 & 3239.98 \\
using\_pc & 28ff9118-e876-497c-9901-cb89a62509fc & 2969.89 \\
using\_pc & 856cedd7-b753-4c4a-bc2e-24f5ca0d6f4b & 1079.99 \\
using\_pc & 9b49246f-30e9-476f-ab8c-56a1bcb1d936 & 1274.97 \\
using\_pc & 9b49246f-30e9-476f-ab8c-56a1bcb1d936 & 1314.97 \\
using\_pc & 52d7e473-06a6-4464-81f4-08199bf5cb6a & 2469.99 \\
using\_pc & f809c446-c3f0-4974-afed-270c12df12a4 & 3224.98 \\
cleaning & a855cacc-0c0f-4ad3-9991-90cbd761637a & 10.07 \\
cleaning & a855cacc-0c0f-4ad3-9991-90cbd761637a & 180.07 \\
cleaning & c0a6b35d-b074-4d99-baa9-79914e4418f7 & 180.08 \\
cleaning & da01e952-d1a1-48ba-9770-88440268ebb9 & 8.08 \\
cleaning & da01e952-d1a1-48ba-9770-88440268ebb9 & 180.08 \\
cleaning & a3f45e46-6c63-402c-a6b9-1c3ad529166d & 2.07 \\
cleaning & 1d7cd66c-4924-4044-af0a-1ce0130c0b60 & 130.07 \\
cleaning & e2f6637c-b3c4-4d19-bef4-0d2dd202f1a6 & 2.07 \\
cleaning & d44e1498-d9ab-4857-ae4b-947d5a4f945a & 30.02 \\
cleaning & a3f45e46-6c63-402c-a6b9-1c3ad529166d & 5.07 \\
eating & 4e554e70-de52-4f24-b5f0-78012b57c5c2 & 1075.99 \\
eating & 50444259-3a59-4f0a-b846-ac0c1b32b5ca & 33.02 \\
eating & 1ce16a97-f614-4660-b21c-4205b33c8bab & 2614.59 \\
eating & 3f2b1d67-50ef-4005-95d3-4cba5bd10b4b & 259.97 \\
eating & 4ff7eacb-814a-4362-9ce2-a439e8cb7786 & 18.07 \\
eating & 4ff7eacb-814a-4362-9ce2-a439e8cb7786 & 170.07 \\
eating & 94e8fb8f-ecf5-409c-940b-a5c966ea2837 & 3341.35 \\
eating & fc6c6313-f1b5-45f7-8272-53096d0d8986 & 180.02 \\
eating & b8276cd7-33e9-4529-80b4-b301b50e227b & 1059.45 \\
eating & 773b6d5f-4ce9-49c4-8aa6-723ec19110f1 & 145.08 \\
\end{longtable}
}
    
\end{appendices}

%%%%%%%%%%%%%%%%%%%%%%%%%%%%%%%%%%%%%%%%%%%%%%%%%%%%%%%%%%%%

\end{document}